\documentclass[11pt]{article}

\usepackage[]{acl}

\usepackage{times}
\usepackage{latexsym}

\usepackage[T1]{fontenc}

\usepackage[utf8]{inputenc}

\usepackage{microtype}

\usepackage{inconsolata}

\usepackage{graphicx}

\usepackage{xspace}
\usepackage{amsmath}
\usepackage{amssymb}
\usepackage{multirow}
\usepackage{booktabs}
\usepackage{makecell}
\usepackage{adjustbox}
\usepackage[linesnumbered,ruled,vlined]{algorithm2e}
\usepackage{bm}
\usepackage{CJKutf8}
\usepackage{tcolorbox}
\usepackage{float}
\usepackage{longtable}
\usepackage{arydshln}

\usepackage[table]{xcolor}
\usepackage{array}
\usepackage{tablefootnote}

\newcommand{\dragon}{\textsc{Loong}\xspace}

\definecolor{headerbg}{HTML}{EAF2FB}
\definecolor{subheaderbg}{HTML}{F6FAFE}
\definecolor{rowgray}{HTML}{FAFAFA}
\definecolor{deltarow}{HTML}{FFF7E6}
\definecolor{dragonrow}{HTML}{EAF7EA}
\definecolor{rulegray}{HTML}{B7C6D9}
\definecolor{bestgreen}{HTML}{1B6E37}

\usepackage{environ}
\NewEnviron{smallequation}{
    \begin{equation}
        \text{\small $\displaystyle \BODY $}
    \end{equation}
}

\title{\dragon: A Human-Like Long Document Translation Agent with Observe-and-Act Adaptive Context Selection}

\author{Yutong Wang$^1$~~
        Xuebo Liu$^1$\thanks{~Xuebo Liu is the corresponding author.}~~
        Derek F. Wong$^2$~~
        Zhilin Li$^1$~~
        Rongqing Jiang$^1$~~\\
        \textbf{Min Zhang}$^3$~~
        \textbf{Shimin Tao}$^3$~~
        \textbf{Daimeng Wei}$^3$~~
        \textbf{Min Zhang}$^1$~~\\
        $^1$Institute of Computing and Intelligence, Harbin Institute of Technology, Shenzhen\\
        $^2$NLP$^2$CT Lab, Department of Computer and Information Science, University of Macau \\ 
        $^3$Huawei Translation Services Center \\
        \texttt{\{wangyutong,lizhilin,jiangtongqing\}@stu.hit.edu.cn}\\
        \texttt{\{liuxuebo,zhangmin2021\}@hit.edu.cn} \;
        \texttt{derekfw@um.edu.com} \\
        \texttt{\{zhangmin186,taoshimin,weidaimeng\}@huawei.com}
}

\begin{document}
\maketitle
\begin{abstract}
Document-level translation remains one of the most challenging tasks for large language models, which are constrained by limited context windows that impede global cohesion, while simultaneously suffering from redundant contextual information that degrades translation quality.
To address this, we propose a human-like long document translation agent called \textbf{\dragon}, which leverages a \textbf{3E} memory module (Essence-Exemplar-Entity) to store summaries, sentence pairs, and entity records as historical context.
Instead of passively attending to all history, \dragon performs deep reasoning to adaptively identify the optimal context for translation guidance.
\dragon optimizes its context policy through reinforcement learning, utilizing preference data derived from its own sampled observe-and-act reasoning trajectories.
Empirical evaluations demonstrate that \dragon achieves substantial translation quality improvements in English $\Leftrightarrow$ Chinese, German, and French directions, with average gains of up to 13.0 points across the three evaluation metrics.
Furthermore, \dragon exhibits strong generalization across domains and robustness against contextual noise, while maintaining remarkable stability in ultra-long document translation.
Our code is released at \url{https://github.com/YutongWang1216/LoongDocMT}.
\end{abstract}

\section{Introduction}

Large language models \citep[LLMs,][]{zhao2026surveylargelanguagemodels} have exhibited exceptional performance across various natural language processing tasks \citep{li2025system, chen2025towards, guan2025deeprag}.
Recently, machine translation (MT) based on prompting LLMs has shown significant advancements in sentence and short-segment translation \citep{peng-etal-2023-towards, he2024exploring, wang-etal-2024-taste, he2025r1, wang2025deep}.
However, in long-document translation scenarios—such as speech transcripts, technical manuals, and literary texts-critical issues arise, including the mismanagement of terminology and inconsistencies in stylistic elements due to their limited contextual capacity \citep{kim2019and, maruf2021survey}.

\begin{figure}
    \centering
    \includegraphics[width=0.8\linewidth]{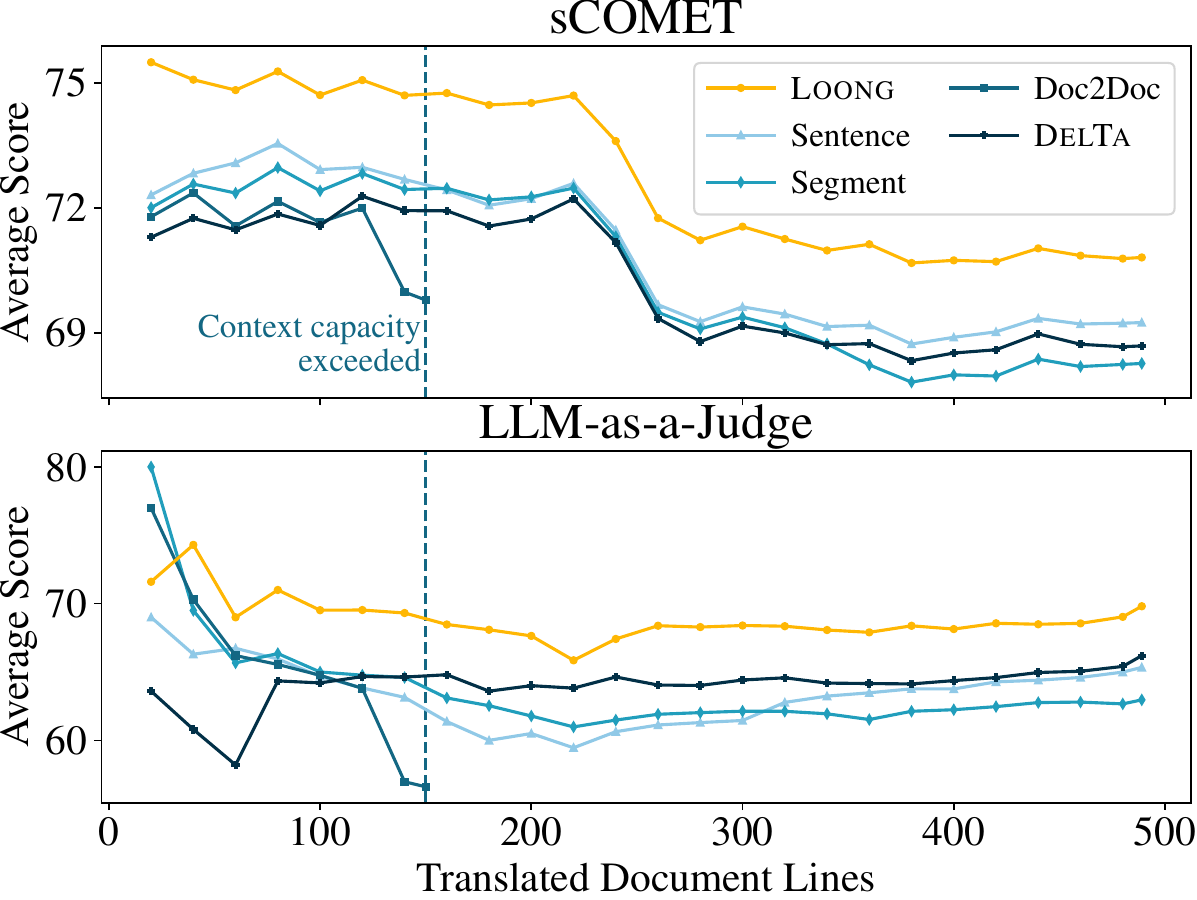}
    \caption{Cumulative average sCOMET and LLM-as-a-Judge scores of \dragon and baseline methods on ultra-long document translation (Chinese $\Rightarrow$ Portuguese). While standard chunking methods (Sentence, Segment), full-history variants \citep[Doc2Doc,][]{wang2023document}, and unfiltered memory agents \citep[\textsc{DelTA},][]{wang2025delta} exhibit continuous degradation or even failure due to context length limits, \dragon successfully distinguishes useful information from retrieved memory to sustain stable, high-quality translations. See \S\ref{sec:analysis} for more experimental details.}
    \label{fig:long_document}
\end{figure}

Document-level machine translation (DocMT) by LLMs has recently attracted increasing attention to address these challenges.
Some studies exploit the extended context windows and long-text processing capabilities of LLMs to generate target documents directly \citep{wang2023document, wu2023exploring, wu2024adapting}.
Other approaches introduce long chain-of-thought (CoT) reasoning or complex multi-step inference procedures to further unlock the DocMT potential of LLMs \citep{dong-etal-2025-two, hu-etal-2025-source, ramos2025multilingual}.
Agent-based workflows \citep{wang-etal-2025-agentdropout, tian-etal-2025-agentinit, wang2026agentdropoutv2optimizinginformationflow, wang2026maspojointpromptoptimization} extract key contextual information to guide the DocMT process \cite{wang2025delta, guo2025doc, li2025tactic, liu2025improving}.
While these methods improve contextual utilization, many of them treat contextual signals indiscriminately, often incorporating extraneous or redundant noise, which ultimately leads to suboptimal performance \citep{karpinska2023large}.

To address these challenges, we introduce \textbf{\dragon}, a human-like long-document translation agent with observe-and-act adaptive context selection.
Mimicking human experts' cognitive processes, \dragon maintains a ``3E'' memory module (Essence-Exemplar-Entity) to store and retrieve segment summaries, bilingual sentence pairs, and entity records as contextual information.
Simultaneously, it employs an observe-and-act reasoning mechanism to distinguish useful context from the retrieved pool, retaining beneficial signals while filtering out redundancy.
We optimize this capability via reinforcement learning (RL) derived from sampled trajectories.
Empirical results demonstrate that \dragon achieves remarkable improvements across diverse backbones (e.g., Qwen, Llama) and language pairs (e.g., English $\Leftrightarrow$ Chinese, German, and French), with average gains of up to 13.0 points on sentence- and document-level metrics.

Our main contributions are as follows:
\begin{itemize}
    \item We develop \dragon, a human-like DocMT agent that retrieves and adaptively identifies useful contextual information by deep reasoning for document translation.
    \item We design a sequential exploration strategy to identify effective context via sampling observe-and-act trajectories, constructing preference data for agent optimization.
    \item We propose an alignment-enforced translation algorithm for document-level methods that ensures the alignment between source and generated target documents.
    \item As shown in Figure~\ref{fig:long_document}, \dragon achieves superior translation quality for ultra-long documents while exhibiting strong generalization across diverse models and domains, as well as robustness against noisy context.
\end{itemize}

\begin{figure*}[t]
    \centering
    \includegraphics[width=0.88\linewidth]{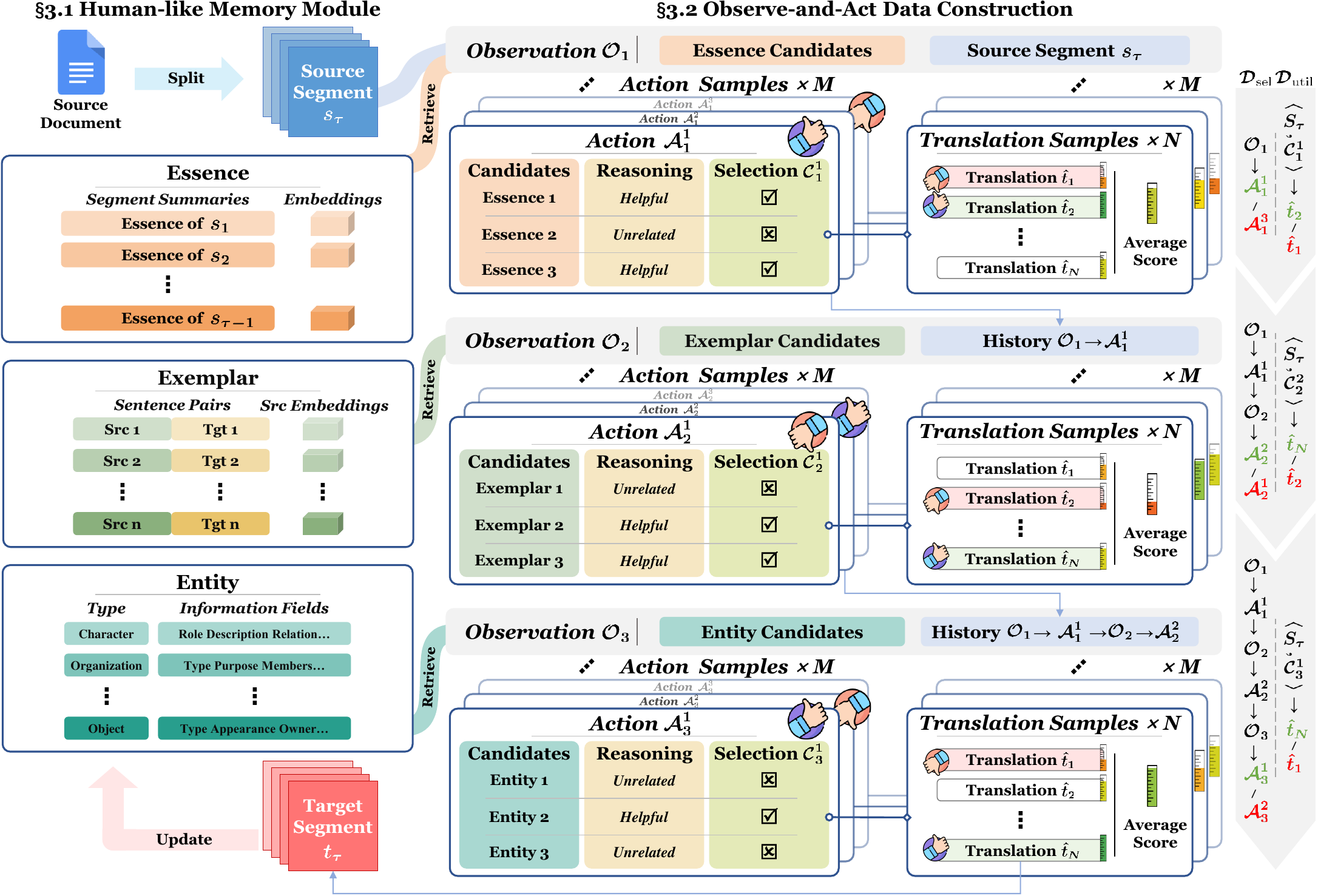}
    \caption{The human-like memory module and observe-and-act dateset construction procedure of \dragon.}
    \label{fig:framework}
\end{figure*}

\section{Related Works}

\paragraph{Document-level MT}
The technical approaches in DocMT can be broadly categorized into two types: Document-to-Sentence (Doc2Sent) and Document-to-Document (Doc2Doc).
\textbf{Doc2Sent} approaches focus on encoding neighboring context to extract discourse features and generate translations for individual sentences.
Advanced neural network designs \citep{wang2017exploiting, tan2021coupling, lyu2021encouraging} have enhanced the modeling of context, enabling more effective sentence-level decoding.
However, Doc2Sent methods still suffer from limited utilization of the target-side context \citep{sun2020rethinking, li2023p}.

In contrast, \textbf{Doc2Doc} methods translate documents segment-by-segment.
Traditional approaches optimize attention allocation in transformers \citep{bao2021g, li2023p}, while LLM-based methods leverage prompting or instruction tuning for high-quality translation \citep{wang2023document, wu2024adapting, liu2025improving}.
However, these methods often fail to maintain strict sentence-level alignment with the source, complicating evaluation and limiting their application in rigorous scenarios \citep{dong-etal-2025-two, wang2025delta}.
To address this, \dragon proposes an LLM-based Doc2Doc agentic workflow, upon which we further optimize the context selection strategies.
We harness the contextual capabilities of LLMs and introduce a specialized algorithm to ensure strict sentence-level alignment between the source and target documents, which combines the strengths of both Doc2Sent and Doc2Doc methods.

\paragraph{Deep Reasoning of MT-LLMs}
In recent years, deep reasoning LLMs such as OpenAI o1 \citep{jaech2024openai} and DeepSeek-R1 \citep{guo2025deepseek} have achieved impressive results on challenging problems such as code generation and question answering \citep{zhang2024o1, guan2025deeprag}.
To further enhance the reasoning and tool-use capabilities of these models, numerous studies focus on sampling their CoT trajectories and providing step-wise rewards to facilitate fine-grained RL \citep{pang2024iterative, li2025iterative, chen2025optima}.

Recent research successfully extends deep reasoning LLMs to MT, demonstrating substantial performance gains \citep{liu2025newtrendsmodernmachine, chen2025evaluatingo1likellmsunlocking, li2025tactic} and tackling open-ended tasks like slang translation \citep{zhao2024marcoo1openreasoningmodels}.
Subsequent works further optimize LLM reasoning for MT, employing RL to develop a ``think-and-translate'' paradigm \citep{he2025r1, wang2025deep} or targeted training for literary nuances \citep{wang-etal-2025-drt}.
However, few of these advancements extend to DocMT.
To overcome the longstanding bottlenecks of document-level issues such as cohesion and coherence, \dragon bridges this gap by introducing an RL-optimized ``observe-and-act'' reasoning process that adaptively filters and integrates multi-granularity context.

\section{Methodology}
We introduce \dragon, a human-like long document translation agent with observe-and-act adaptive context selection.
Drawing inspiration from professional human translators, we design a multi-granularity contextual memory module that records and retrieves historical information during translation (\S\ref{sec:memory_module}).
To filter out redundant information from the retrieved, we implement a deep reasoning mechanism to identify relevant items, and explore optimal context strategies by sampling diverse reasoning trajectories (\S\ref{sec:deep_reasoning}).
Based on this, we apply RL to optimize the agent's policy, and deploy it alongside an alignment-enforced translation algorithm, yielding high-quality and strictly aligned target documents (\S\ref{sec:reinforcement_learning}).

\subsection{Human-like Translation Memory Module}\label{sec:memory_module}

Human translators typically follow an integrated reading--translation workflow when handling long documents: they translate the text segment by segment while continuously memorizing critical contextual information to maintain coherence and cohesion.
To emulate this cognitive process, we partition the source document into a sequence of $L$ segments $\{\bm{s}_1, \ldots, \bm{s}_L\}$, where each segment contains $l$ sentences.
Similar to \citet{wang2025delta}, we design a multi-granularity memory architecture, designated as the \textbf{3E Module}, which is updated after completing each segment. 
As shown in the left part of Figure \ref{fig:framework}, the module synergizes three distinct components to capture information at different granularities: \textit{Essence} (Global), \textit{Exemplars} (Pattern), and \textit{Entities} (Specific).

\paragraph{Essence}
To capture semantic progression, we maintain a memory bank $\mathcal{E}_\text{s}^{(\tau-1)}$ storing LLM-generated summaries of the preceding $\tau-1$ segments.
Before translating segment $\bm{s}_{\tau}$, we use a sentence encoder to extract embeddings and retrieve a candidate subset $\tilde{\mathcal{E}}_\text{s}^{(\tau-1)}$ containing the top-$K_\text{s}$ summaries with the highest cosine similarity to $\bm{s}_{\tau}$ for the agent's further selection.

\paragraph{Exemplars}
To maintain stylistic consistency, an exemplar memory $\mathcal{E}_\text{x}^{(\tau-1)}$ stores all previously processed source-target sentence pairs as in-context demonstrations.
Similar to Essence, we retrieve a subset $\tilde{\mathcal{E}}_\text{x}^{(\tau-1)}$ of the top-$K_\text{x}$ exemplars based on the source-sentence embedding similarity to $\bm{s}_{\tau}$, providing stylistic references to guide the translation.

\paragraph{Entities}
To enforce terminological precision, we maintain a structured knowledge base $\mathcal{E}_\text{n}^{(\tau-1)}$, where each entry $(e^\text{src}, e^\text{tgt}, \mathcal{I})$ records an entity's bilingual names and domain-specific attributes.
After translating each segment, the agent extracts entities, classifies them, and updates their attributes $\mathcal{I}$ (see Appendix \ref{sec:app_filed} for more details).
For $\bm{s}_{\tau}$, we retrieve all mentioned entities and generate their context-aware descriptions $\bm{d}_i$ conditioned on $\bm{s}_{\tau}$, forming the candidate knowledge set: $\tilde{\mathcal{E}}_\text{n}^{(\tau)} = \{(e_i^\text{src}, e_i^\text{tgt}, \bm{d}_i) \mid e_i^\text{src} \in \bm{s}_{\tau}\}$.

\subsection{Observe-and-Act Data Construction}\label{sec:deep_reasoning}

To enable the agent to accurately identify and effectively utilize beneficial context for translation, we focus on exploring diverse context reasoning strategies and collecting preference data.
However, performing a joint search over all memory types results in an intractable combinatorial space.
Consequently, we decompose the exploration process into three sequential steps, where each step exclusively investigates the utility of one specific context type.
This sequential design is driven by two critical advantages.
Functionally, it transforms the search space from a combinatorial product ($O(\prod{2^K})$) to an additive sum ($O(\sum{2^K})$), rendering the trajectory exploration feasible.
Strategically, it allows us to obtain fine-grained feedback on the specific contribution of each context type, facilitating precise adjustments to the agent's strategy.

\paragraph{Observe-and-Act Reasoning Process}
Based on this decoupled framework, we aim to construct preference datasets targeting two complementary capabilities: \textit{context selection} (determining what to select) and \textit{context utilization} (determining how to translate given the context).
As shown in the right part of Figure \ref{fig:framework}, we implement this framework via a multi-step \textbf{observe-and-act} workflow.
For a specific segment $\bm{s}_{\tau}$, we organize the candidate contextual information retrieved from the previous $\tau-1$ segments (\S\ref{sec:memory_module}) into an ordered sequence $\mathbf{E} = \langle\tilde{\mathcal{E}}_\text{s}^{(\tau-1)}, \tilde{\mathcal{E}}_\text{x}^{(\tau-1)}, \tilde{\mathcal{E}}_\text{n}^{(\tau-1)}\rangle$.
The agent then executes a 3-step reasoning $\left\langle \mathcal{O}_1, \mathcal{A}_1, \mathcal{O}_2, \mathcal{A}_2, \mathcal{O}_3, \mathcal{A}_3 \right\rangle$, where the first step $(\mathcal{O}_1, \mathcal{A}_1)$ is dedicated to observing and selecting from the Essence candidates, followed sequentially by the processing of Exemplars (Step 2) and Entities (Step 3).
At step $k$, the observation $\mathcal{O}_k = \left\{\mathcal{H}_{k-1}, \mathbf{E}[k]\right\}$ encapsulates the history of all previous reasoning steps $\mathcal{H}_{k-1}$ and the specific set of retrieved candidates $\mathbf{E}[k]$.
The corresponding action $\mathcal{A}_k$ is structured as a composite output $\langle \bm{r}_k, \mathcal{C}_k \rangle$: a \textbf{reasoning thought} $\bm{r}_k$ that explicitly analyzes the semantic relevance of each item in $\mathbf{E}[k]$ with respect to $\bm{s}_{\tau}$, followed by the \textbf{final selection} $\mathcal{C}_k$, a subset of items deemed most beneficial for the translation process.

\paragraph{Context Selection Strategy Sampling}
To explore diverse context selection strategies, we perform parallel sampling on $\mathcal{A}_k$ for $M$ times:
\begin{equation}
    \mathcal{G}_k = \left\{ \mathcal{A}_k^{i} \right\}_{i=1}^M, \text{where } \mathcal{A}_k^{i} \sim \bm{\pi}_\theta(\cdot\mid\mathcal{O}_k).
\end{equation}
Based on each candidate action, the agent translates $\bm{s}_{\tau}$ using the selected context items as additional guidance.
We sample $N$ translations given each candidate action $\mathcal{A}_k^{i} = \langle r_k^{i}, \mathcal{C}_k^{i} \rangle$:
\begin{equation}
    \mathcal{T}_k^{i} = \left\{ \hat{\bm{t}}_j \right\}_{j=1}^N, \hat{\bm{t}}_j \sim \bm{\pi}_\theta(\cdot \mid \bm{s}_{\tau}, \mathcal{C}_k^{i}).
\end{equation}
We calculate the average quality score of $\mathcal{T}_k^{i}$ as the utility proxy for $\mathcal{A}_k^{i}$:
\begin{equation}
    U(\mathcal{A}_k^{i}) = \frac{1}{N}\sum_{\hat{\bm{t}} \in \mathcal{T}_k^{i}}\mu(\bm{s}_{\tau}, \hat{\bm{t}}, \bm{t}^*).
\end{equation}
where $\mu(\cdot)$ denotes the evaluation metric and $\bm{t}^*$ is the reference translation.
Accordingly, we identify the action yielding the highest utility as the \textit{preferred} strategy $\mathcal{A}_k^+$, and the lowest as the \textit{dispreferred} $\mathcal{A}_k^-$:
\begin{equation}
    \mathcal{A}_k^+ = \operatorname*{argmax}_{\mathcal{A} \in \mathcal{G}_k} U(\mathcal{A}),\ 
    \mathcal{A}_k^- = \operatorname*{argmin}_{\mathcal{A} \in \mathcal{G}_k} U(\mathcal{A}).
\end{equation}
The triplet $(\mathcal{O}_k, \mathcal{A}_k^+, \mathcal{A}_k^-)$ constitutes a sample in the \textbf{context selection strategy dataset} $\mathcal{D}_\text{sel}$.

\paragraph{Context Utilization Strategy Sampling}
To further explore the diverse context utilization strategies to enhance the translation process, we select the highest- and lowest-scoring translations in $\mathcal{T}_k^{i}$:
\begin{align}
    \bm{t}^+ &= \operatorname*{arg\,max}_{\hat{\bm{t}} \in \mathcal{T}_k^{i}} \ \mu(\bm{s}_\tau, \hat{\bm{t}}, \bm{t}^*), \\
    \bm{t}^- &= \operatorname*{arg\,min}_{\hat{\bm{t}} \in \mathcal{T}_k^{i}} \ \mu(\bm{s}_\tau, \hat{\bm{t}}, \bm{t}^*).
\end{align}
We then combine the input $\langle \bm{s}_{\tau},\mathcal{C}_k^{i} \rangle$ with this pair to form a triplet $(\langle \bm{s}_{\tau},\mathcal{C}_k^{i} \rangle, \bm{t}^+, \bm{t}^-)$ in the \textbf{context utilization strategy dataset} $\mathcal{D}_\text{util}$.

The optimal action $\mathcal{A}_k^+$ are incorporated into the observation for the next step $\mathcal{O}_{k+1}$, and the optimal translation $\bm{t}^+$ derived in the last step is established as the final translation for $\bm{s}_{\tau}$, which is then used to update the memory module following the procedure in \S\ref{sec:memory_module}.
Finally, we merge the datasets into the overall training set $\mathcal{D} = \mathcal{D}_\text{sel} \cup \mathcal{D}_\text{util}$.
The pseudo code of the preference data construction is shown in Appendix \ref{app:code}.

\subsection{Agent Tuning \& Inference}\label{sec:reinforcement_learning}

\paragraph{Preference Tuning}
We utilize the constructed preference data to optimize the agent's context strategies.
To equip the LLM with the initial ability to perform deep reasoning and output results, we first perform a cold-start supervised finetuning (SFT), using only the preferred instances from $\mathcal{D}$.
The stage maximizes the following objective:
\begin{equation}
    \mathbb{E}_{(\bm{x}, \bm{y}^+) \sim \mathcal{D}} \left[ \log \bm{\pi}_\theta\left(\bm{y}^+ \mid \bm{x}\right) \right],
\end{equation}
where $\theta$ denotes the trainable parameters of the original LLM.
Next, we apply direct preference optimization (DPO) to further refine the LLM's context-selection and translation strategies using the full preference dataset.
The DPO objective is
\begin{equation}
    \mathbb{E}_{(\bm{x}, \bm{y}^+, \bm{y}^-) \sim \mathcal{D}} \left[ \log \sigma\left( \beta \Delta(\theta, \theta_{\text{SFT}}) \right) \right],
\end{equation}
where  $\theta$ and $\theta_\text{SFT}$ denote the parameters being updated during DPO and those obtained after SFT, respectively.
The advantage difference $\Delta(\theta, \theta_{\text{SFT}})$ between the preferred and dispreferred outputs is defined as follows:
\begin{equation}
    \Delta = \log \frac{\bm{\pi}_\theta(\bm{y}^+ \mid \bm{x})}{\bm{\pi}_{\theta_{\text{SFT}}}(\bm{y}^+ \mid \bm{x})} - \log \frac{\bm{\pi}_\theta(\bm{y}^- \mid \bm{x})}{\bm{\pi}_{\theta_{\text{SFT}}}(\bm{y}^- \mid \bm{x})}.
\end{equation}
To reduce GPU memory consumption and improve training efficiency, we employ Low-Rank Adaptation \citep[LoRA,][]{hu2022lora} during DPO.

\paragraph{Alignment-Enforced Inference Procedure}
During inference, the trained agent samples only once for the selection of each type of contextual information.
No intermediate translations or quality assessments are generated at this stage.
After all three reasoning steps are executed, the agent produces a final translation of the current segment guided by the selected contextual items, and this translation is subsequently used to update the memory module following the same process as \S\ref{sec:memory_module}.
Crucially, to ensure this generation maintains strict sentence-level alignment with the source segments for accurate evaluation and memory updates, we employ a recursive alignment enforcement algorithm.
For a segment $\bm{u}_{i:j}$ to be translated, we inject sentence indices and separators, which the LLM is told to preserve.
Once the output failed to align with the source sentences, the segment is recursively split into halves ($\bm{u}_{i:k}$ and $\bm{u}_{k+1:j}$, where $k=i-1+\lfloor \frac{j-i+1}{2} \rfloor$) until alignment is achieved or the segment reduces to a single sentence:
\begin{equation}
    \operatorname{T}(\bm{u}_{i:j}) = 
    \begin{cases}
       \operatorname{LLM}(u_{i:j}), \text{if aligned or } i=j & \\
       \operatorname{T}(\bm{u}_{i:k}) \oplus \operatorname{T}(\bm{p}_{k+1:j}), \text{otherwise}&
    \end{cases}
\end{equation}
This guarantees an aligned translation.
Algorithm details are shown in Appendix \ref{app:code}.
For a complete case of \dragon, please refer to Appendix \ref{app:case_study}

\begin{table}[t]
    \centering
    \scalebox{0.65}{
    \renewcommand{\arraystretch}{1.1}
    \begin{tabular}{>{\raggedright\arraybackslash}p{5.5cm}rl}
        \toprule
        \textbf{Dataset} & \textbf{Words / Doc.} & \textbf{Domain} \\
        \midrule
        \multicolumn{3}{l}{\textit{\textbf{In-Domain}}} \\
        \hdashline
        News Commentary V18.1\tablefootnote{https://data.statmt.org/news-commentary/v18.1} & 980 & News \\
        WMT24++ \citep{deutsch2025wmt24expandinglanguagecoverage}\tablefootnote{https://huggingface.co/datasets/google/wmt24pp} & 637 & News, etc. \\
        \midrule
        \multicolumn{3}{l}{\textit{\textbf{Out-of-Domain}}} \\
        \hdashline
        News Commentary (Unseen Langs) & 980 & News \\
        IWSLT2017 \citep{cettolo-etal-2017-overview}\tablefootnote{https://huggingface.co/datasets/IWSLT/iwslt2017} & 1,939 & TED Talks \\
        GuofengV1 \citep{wang2023findings}\tablefootnote{https://github.com/longyuewangdcu/GuoFeng-Webnovel} & 1,445 & Webnovels \\
        Journey to the West & 51,854 & Ultra-Long Novel \\
        \bottomrule
    \end{tabular}}
    \caption{Summary of the evaluation datasets.}
    \label{tab:evaluation_datasets}
\end{table}

\begin{table*}[t]
    \centering
    \renewcommand{\arraystretch}{1.1}
    \setlength{\tabcolsep}{4.5pt}
    \resizebox{\textwidth}{!}{
        \begin{tabular}{cl cccc cccc cccc cccc}
            \toprule
            \multirow{3.6}{*}{Model} & \multirow{3.6}{*}{System} & \multicolumn{8}{c}{News Commentary V18.1} & \multicolumn{8}{c}{WMT24++} \\
            \cmidrule(lr){3-10} \cmidrule(lr){11-18}
             & & \multicolumn{4}{c}{En $\Rightarrow$ Xx} & \multicolumn{4}{c}{Xx $\Rightarrow$ En} & \multicolumn{4}{c}{En $\Rightarrow$ Xx} & \multicolumn{4}{c}{Xx $\Rightarrow$ En} \\
            \cmidrule(lr){3-6} \cmidrule(lr){7-10} \cmidrule(lr){11-14} \cmidrule(lr){15-18}
             & & sCMT & dCMT & LLM & Avg & sCMT & dCMT & LLM & Avg & sCMT & dCMT & LLM & Avg & sCMT & dCMT & LLM & Avg \\
            \midrule
            \multirow{5}{*}{\rotatebox{90}{Qwen2.5-7B}}
            & Sentence          & 84.4 & 79.1 & 56.2 & 73.2 & 86.7 & 82.1 & 78.4 & 82.4 & 74.3 & 71.1 & 44.9 & 63.4 & 77.8 & 74.6 & 57.3 & 69.9 \\
            & Segment           & 84.9 & 79.6 & 64.0 & 76.2 & 86.5 & 80.2 & 79.6 & 82.1 & 79.6 & 77.0 & 62.7 & 73.1 & 82.7 & 78.2 & 73.8 & 78.2 \\
            & Doc2Doc           & 84.5 & 79.1 & 63.1 & 75.6 & 85.7 & 78.3 & 74.3 & 79.5 & 80.1 & 77.4 & 64.9 & 74.1 & 83.0 & 78.8 & 74.8 & 78.8 \\
            & \textsc{DelTA}    & 85.1 & 79.8 & 63.4 & 76.1 & 87.4 & 83.1 & 87.6 & 86.0 & 80.0 & 77.5 & \textbf{65.2} & 74.3 & 83.9 & 81.4 & 80.3 & 81.9 \\
            & \dragon           & \textbf{85.8} & \textbf{80.5} & \textbf{66.7} & \textbf{77.7} & \textbf{88.0} & \textbf{83.7} & \textbf{90.7} & \textbf{87.5} & \textbf{81.1} & \textbf{78.7} & 65.0 & \textbf{74.9} & \textbf{84.4} & \textbf{82.1} & \textbf{82.0} & \textbf{82.9} \\
            \midrule
            \multirow{5}{*}{\rotatebox{90}{Qwen3-8B}}
            & Sentence          & 86.9 & 81.6 & 78.9 & 82.5 & 87.6 & 83.1 & 87.8 & 86.2 & 81.9 & 79.4 & 72.5 & 77.9 & 83.5 & 80.9 & 75.8 & 80.0 \\
            & Segment           & 86.5 & 80.9 & 78.3 & 81.9 & 87.2 & 81.8 & 82.6 & 83.9 & 81.1 & 77.4 & 70.9 & 76.5 & 83.5 & 79.3 & 74.2 & 79.0 \\
            & Doc2Doc           & 86.4 & 80.7 & 79.3 & 82.1 & 87.6 & 82.6 & 86.2 & 85.4 & 82.0 & 79.0 & 74.4 & 78.5 & 83.8 & 80.4 & 78.6 & 80.9 \\
            & \textsc{DelTA}    & 86.7 & 81.4 & 78.8 & 82.3 & 87.7 & 83.4 & 89.1 & 86.7 & 81.8 & 79.5 & 75.1 & 78.8 & 84.2 & 81.9 & 81.1 & 82.4 \\
            & \dragon           & \textbf{87.3} & \textbf{82.1} & \textbf{81.7} & \textbf{83.7} & \textbf{88.1} & \textbf{83.7} & \textbf{90.2} & \textbf{87.3} & \textbf{82.8} & \textbf{80.4} & \textbf{76.8} & \textbf{80.0} & \textbf{84.8} & \textbf{82.5} & \textbf{83.5} & \textbf{83.6} \\
            \midrule
            \multirow{5}{*}{\rotatebox{90}{Qwen3-14B}}
            & Sentence          & 87.4 & 82.2 & 84.4 & 84.7 & 87.9 & 83.5 & 88.2 & 86.5 & 82.8 & 80.5 & 79.2 & 80.8 & 84.3 & 81.8 & 80.4 & 82.2 \\
            & Segment           & 87.3 & 82.2 & 83.5 & 84.4 & 88.0 & 83.6 & 89.1 & 86.9 & 82.9 & 79.8 & 78.8 & 80.5 & 84.8 & 82.5 & 82.6 & 83.3 \\
            & Doc2Doc           & 87.4 & 82.2 & 84.8 & 84.8 & 88.0 & 83.8 & 90.0 & 87.3 & 83.2 & 80.5 & 80.5 & 81.4 & 84.9 & 82.6 & 85.1 & 84.2 \\
            & \textsc{DelTA}    & 87.5 & 82.3 & 85.0 & 84.9 & 88.0 & 83.7 & 90.4 & 87.4 & 82.9 & 80.5 & \textbf{80.7} & 81.4 & 84.7 & 82.4 & 84.3 & 83.8 \\
            & \dragon           & \textbf{87.7} & \textbf{82.6} & \textbf{85.6} & \textbf{85.3} & \textbf{88.2} & \textbf{84.0} & \textbf{91.0} & \textbf{87.7} & \textbf{83.4} & \textbf{80.8} & 80.5 & \textbf{81.6} & \textbf{85.0} & \textbf{82.7} & \textbf{85.3} & \textbf{84.4} \\
            \midrule
            \multirow{5}{*}{\rotatebox{90}{Llama3.1-8B}}
            & Sentence          & 84.8 & 78.6 & 65.4 & 76.3 & 86.8 & 82.3 & 81.9 & 83.7 & 76.7 & 73.7 & 54.7 & 68.4 & 80.5 & 77.7 & 69.6 & 75.9 \\
            & Segment           & 84.3 & 78.3 & 61.1 & 74.6 & 86.9 & 82.4 & 82.0 & 83.8 & 77.9 & 73.9 & 55.7 & 69.2 & 83.6 & 80.8 & 74.6 & 79.6 \\
            & Doc2Doc           & 84.6 & 79.1 & 64.8 & 76.2 & 86.9 & 82.2 & 82.2 & 83.7 & 78.9 & 77.0 & 59.9 & 71.9 & 83.6 & 81.0 & 76.7 & 80.4 \\
            & \textsc{DelTA}    & 85.5 & 79.8 & 71.2 & 78.8 & 86.8 & 82.3 & 84.8 & 84.7 & 78.5 & 76.3 & 62.1 & 72.3 & 79.7 & 76.7 & 75.2 & 77.2 \\
            & \dragon           & \textbf{86.3} & \textbf{81.3} & \textbf{73.0} & \textbf{80.2} & \textbf{87.8} & \textbf{83.5} & \textbf{87.6} & \textbf{86.3} & \textbf{80.4} & \textbf{77.9} & \textbf{64.4} & \textbf{74.3} & \textbf{84.1} & \textbf{81.7} & \textbf{82.3} & \textbf{82.7} \\
            \bottomrule
        \end{tabular}
    }
    \caption{Main results of \dragon compared to the baseline methods. ``sCMT'' and ``dCMT'' denote sCOMET and dCOMET, respectively. The best results are marked in \textbf{bold font}. All models are instruction-tuned versions.}
    \label{tab:main_results}
\end{table*}

\section{Experiments}

\subsection{Setup}
\paragraph{Datasets}

To align with our focus on long-document translation, we construct the training set by selecting documents comprising more than 50 lines from the News Commentary V18.1\footnote{https://data.statmt.org/news-commentary/v18.1} dataset.
From this filtered pool, we randomly sample approximately 500 documents for each language pair of English (En) $\Leftrightarrow$ Chinese (Zh), German (De), and French (Fr).
For evaluation in both the main experiments and subsequent analyses, we assess our approach across a diverse set of in-domain and out-of-domain benchmarks.
Table~\ref{tab:evaluation_datasets} summarizes the datasets utilized for evaluation, detailing their average document lengths and respective domains.
The in-domain News Commentary test set is drawn exclusively from the held-out portion of the data not utilized during training.
Similar to the training data, we select longer subsets for both the News Commentary and WMT24++ test sets.
Detailed information of these datasets and data preprocessing methods (i.e., length filtering and sampling) is provided in Appendix~\ref{app:dataset}.

\paragraph{Metrics}
We adopt sCOMET \citep{rei-etal-2022-comet}\footnote{https://github.com/Unbabel/COMET} as the evaluation metric $\mu$ during data construction in \S\ref{sec:deep_reasoning}, and employ sCOMET and dCOMET \citep{easy_doc_mt}\footnote{https://github.com/amazon-science/doc-mt-metrics} as the sentence- and document-level translation quality metrics, respectively, both instantiated with the wmt22-comet-da model.
Additionally, we use GPT-4.1 to evaluate the translated documents across five dimensions:
General Quality, Cohesion, Coherence, Style Consistency, and Terminology Consistency.
Each of these five dimensions is scored on a scale from 1 to 100, with the final meta-score being the average of the individual scores.
The metric design prompt is provided in Appendix~\ref{app:prompt}.

\paragraph{Models}
We employ all-distilroberta-v1\footnote{https://huggingface.co/sentence-transformers/all-distilroberta-v1} as the sentence encoder, and adopt four open-source LLMs as the backbone models: Qwen2.5-7B-Instruct \citep{qwen2.5}\footnote{https://huggingface.co/Qwen/Qwen2.5-7B-Instruct}, Qwen3-8B \citep{qwen3technicalreport}, Qwen3-14B, and Llama3.1-8B-Instruct \citep{grattafiori2024llama}\footnote{https://huggingface.co/meta-llama/Llama-3.1-8B-Instruct}.
For each model, we train two variants: an \textbf{En $\Rightarrow$ Xx} model using training data from En $\Rightarrow$ Zh, De, Fr directions, and an \textbf{Xx $\Rightarrow$ En} model using data from the opposite directions.
Each model is trained on 4 A800 GPUs, consuming around 20 hours for the whole training process.

\paragraph{Hyperparameters}
We set the segment length $l$ to 5 sentences.
In the memory module, both the summary and exemplar retrieval sizes, $K_{\text{s}}$ and $K_{\text{x}}$, are set to 4.
For preference data construction, we set the action sampling count $M$ to 7, and the translation sampling count $N$ to 5.
During SFT, we train the models for 1 epoch with a batch size of 64 and a learning rate of 1e-5, using DeepSpeed ZeRO-Stage 3 optimization \citep{rajbhandari2020zero}.
During DPO, we train the models for 1 epoch with a batch size of 32 and a learning rate of 5e-6.
We set $\beta = 0.1$ and LoRA rank to 8.
The maximum text length for both stages is 2,560.
For both preference data generation and test-time inference, we use a temperature of 0.7 and a top-$p$ value of 1.0.

\paragraph{Baselines}
We compare \dragon against the following baseline approaches:
a) Sentence: A vanilla sentence-level baseline that translates documents in an isolated, sentence-by-sentence manner.
b) Segment: A simple document-level baseline that translates documents segment by segment without leveraging across-segment context.
c) Doc2Doc \citep{wang2023document}: A DocMT approach in which the LLM translates one segment per conversation turn while storing all previously translated segments in the conversation history.
d) \textsc{DelTA} \citep{wang2025delta}: A DocMT agent equipped with a multi-granularity memory module that stores and retrieves key contextual information to support context-aware translation without further filtering.

\subsection{Main Results}

As shown in Table \ref{tab:main_results}, \dragon consistently achieves state-of-the-art overall performance across both the News Commentary V18.1 and WMT24++ test sets.
Regardless of the backbone model employed or the translation direction, our method attains the highest average scores.
Detailed results for individual language pairs are provided in Appendix \ref{app:results}.

\begin{table}[t]
    \centering
    \scalebox{0.65}{
    \begin{tabular}{l|lcccc}
        \toprule
        Id & Method & sCMT & dCMT & LLM & Avg \\
        \midrule
        1 & \dragon           & \textbf{86.4} & \textbf{81.3} & \textbf{73.0} & \textbf{80.2} \\
        \hdashline
        2 & w/o Context       & 85.9 & 80.6 & 65.6 & 77.4 \\
        3 & w/o Translation   & 79.3 & 65.4 & 46.1 & 63.6 \\
        4 & w/o Tuning        & 84.6 & 79.0 & 62.5 & 75.4 \\
        \hdashline
        5 & w/o Essence       & 86.2 & 81.1 & 69.7 & 79.0 \\
        6 & w/o Exemplar      & 86.3 & 81.1 & 70.6 & 79.3 \\
        7 & w/o Entity        & 86.2 & 81.0 & 72.0 & 79.7 \\
        \bottomrule
    \end{tabular}}
    \caption{Results of the ablation study. LLama3.1-8B-Instruct is adopted as the backbone model. ``sCMT'' and ``dCMT'' denote sCOMET and dCOMET, respectively. We adopt LLama3.1-8B-Instruct trained in En $\Rightarrow$ Xx directions for the ablation experiments.}
    \label{tab:ablation}
\end{table}

\paragraph{Sentence-Level Quality Improvement}
Incorporating a broad context risks introducing noise that degrades sentence-level accuracy.
For example, using the Qwen3-8B backbone on News Commentary (En $\Rightarrow$ Xx), document-level baselines (\textsc{DelTA} and Doc2Doc) yield sCOMET scores (86.7 and 86.4) that actually fall behind the naive Sentence baseline (86.9).
Overcoming this, \dragon effectively filters irrelevant memory, achieving the highest score of 87.3.
\dragon's consistent top performance across all datasets confirms the robust efficacy of our deep reasoning mechanism across varying benchmarks.

\paragraph{Document-Level Quality Improvement}
\dragon consistently dominates context-aware dCOMET and multi-dimensional LLM metrics.
Notably, using Llama3.1-8B-Instruct on WMT24++ (En $\Rightarrow$ Xx), it attains a dCOMET of 77.9, surpassing the Sentence baseline by 4.2 points.
More impressively, in the LLM-as-a-Judge evaluation assessing overall discourse fluency and consistency, \dragon achieves an 82.3 in Xx $\Rightarrow$ En, outperforming the state-of-the-art \textsc{DelTA} (75.2) by a remarkable 7.1 points.
This substantial gap highlights our agent's superior capability in modeling long-range dependencies and resolving discourse coherence issues.

\paragraph{Robustness Across Model Families and Sizes}
\dragon consistently performs well across diverse backbones (Qwen2.5/3, Llama3.1) and sizes ranging from 7B to 14B.
Notably, it achieves a peak average LLM score of 91.0 on Qwen3-14B.
This wide-ranging consistency validates our agent as a robust, model-agnostic framework regardless of the underlying LLM's architecture or scale.

\subsection{Ablation Study}

\paragraph{Training Configuration Ablation}
To disentangle the contribution of the optimized context strategy from simple exposure to translation targets, we train a variant exclusively on translation data, bypassing the observe-and-act reasoning process (Line 2 in Table \ref{tab:ablation}).
The resulting performance drop confirms that translation supervision alone is insufficient, and our gains heavily rely on the learned context strategies.
We further evaluate variants trained solely on context selection preferences (Line 3) or without any tuning (Line 4).
Consistent degradation across all these settings demonstrates that the synergy of context selection and utilization data is indispensable for the agent to master effective context strategies for high-quality translation.

\begin{figure}
    \centering
    \includegraphics[width=0.75\linewidth]{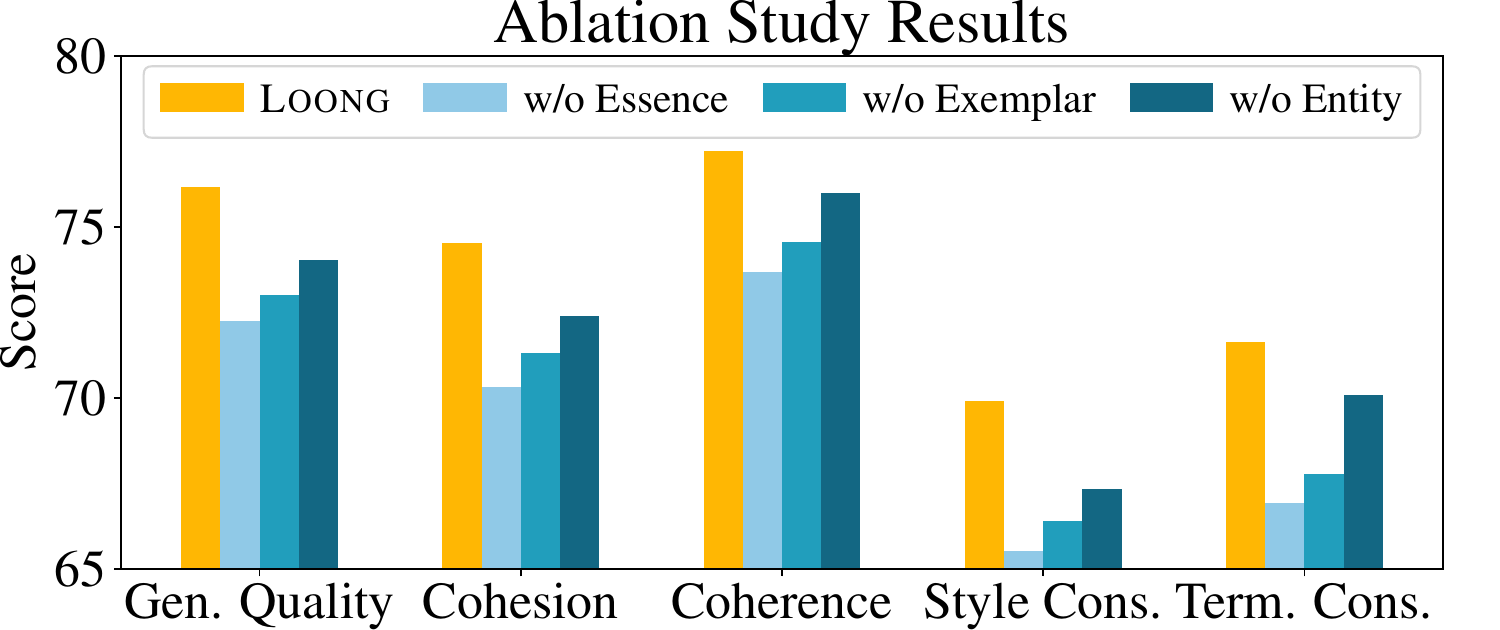}
    \caption{Performance changes across individual LLM evaluation dimensions after removing each memory module component.}
    \label{fig:llm_dimension}
\end{figure}

\paragraph{Memory Component Ablation}

To validate the necessity of each memory type, we individually exclude Essence, Exemplars, and Entities information from the training trajectories.
Results in Lines 5–7 of Table \ref{tab:ablation} and Figure \ref{fig:llm_dimension} reveal consistent degradation upon removing any component, indicating their collective importance.
Notably, eliminating Essence causes the most severe drop.
We hypothesize that global summaries provide critical content and genre signals, making them paramount for maintaining coherence across long documents.

\subsection{Analysis}\label{sec:analysis}

\begin{figure}[t]
    \centering
    \includegraphics[width=0.9\linewidth]{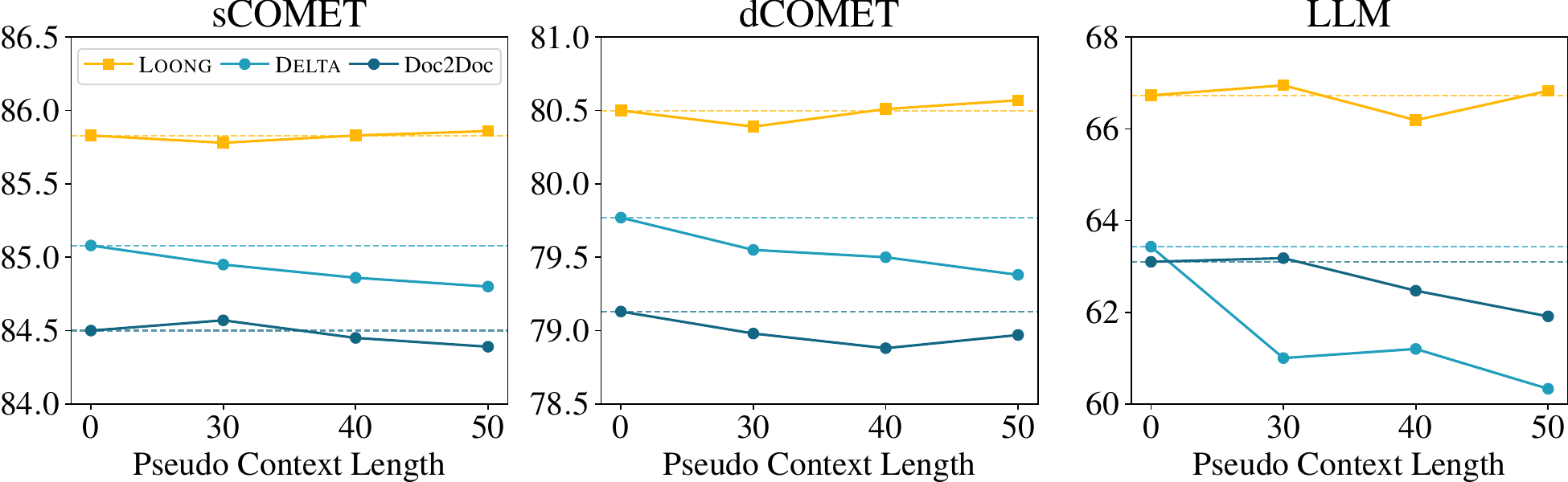}
    \caption{Translation performance on test sets augmented with pseudo-context. A pseudo-context length of 0 indicates the original results without disturbance.}
    \label{fig:robust}
\end{figure}

\paragraph{Robustness on Context}
To evaluate \dragon's robustness against context noises, we inject distracting pseudo-context (30-50 sentences sampled from other documents) into the source.
We compare our results achieved by Qwen2.5 against the selection-free baselines in En $\Rightarrow$ Xx directions.
As shown in Figure \ref{fig:robust}, two baselines suffer consistent degradation simultaneously on sCOMET, dCOMET, and LLM scores as noise increases, whereas \dragon remains largely stable, indicating its robustness against context disturbance.
Interestingly, \dragon sometimes achieves marginal gains, likely by selectively exploiting domain signals within the pseudo-context while successfully filtering out noise.

\begin{figure}
    \centering
    \includegraphics[width=0.75\linewidth]{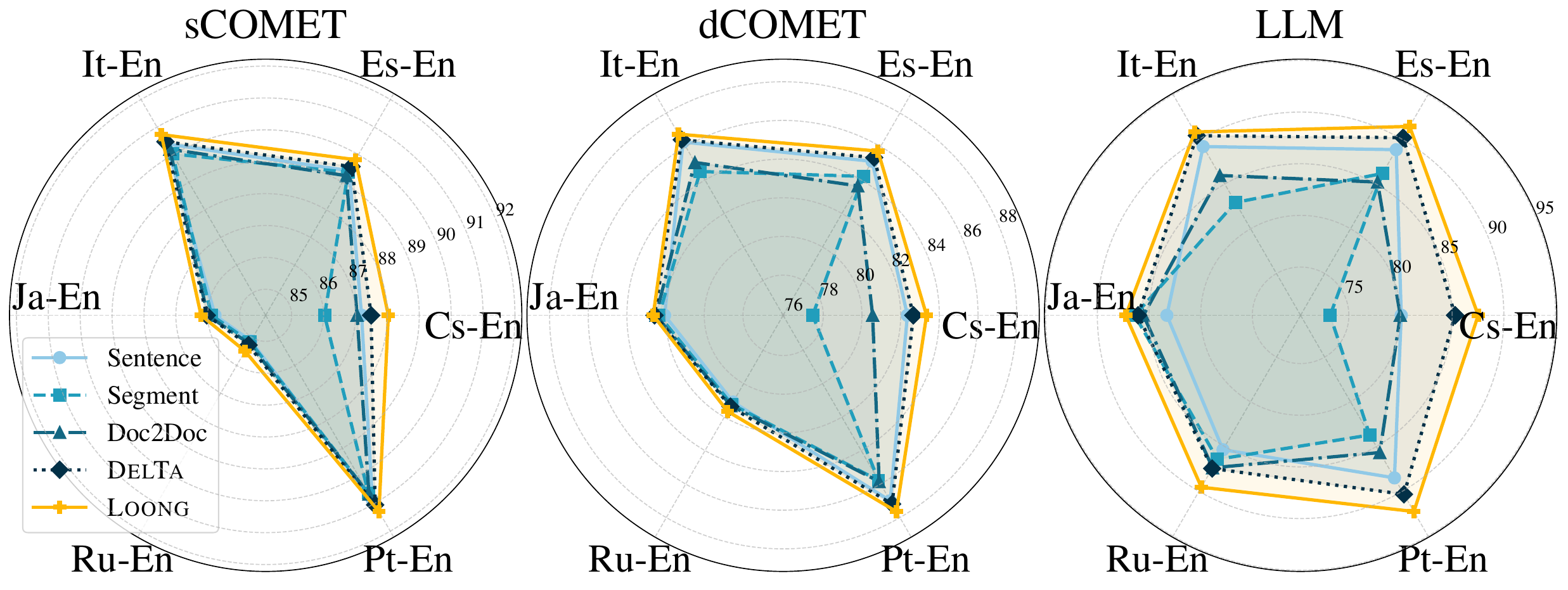}
    \caption{Test results on the unseen language pairs.}
    \label{fig:lang_ood}
\end{figure}

\paragraph{Transferability between Languages}
We assess the cross-lingual transferability of \dragon by evaluating its translation performance of the XX $\Rightarrow$ En version with Qwen3-8B on the unseen Czech (Cs), Spanish (Es), Italian (It), Japanese (Ja), Russian (Ru), and Portuguese (Pt) $\Rightarrow$ English test sets sampled from News Commentary.
As shown in Figure \ref{fig:lang_ood}, \dragon still achieves consistent performance improvements on all these language pairs, suggesting that the context selection and utilization strategy learned in some languages can be generalized to other language pairs, rendering \dragon applicable to languages with scarce training data.

\paragraph{Generalization across Domains}
To explore the generalization capability of \dragon on out-of-domain (OOD) data, we evaluate the En $\Rightarrow$ Xx model with Llama3.1-8B-Instruct as backbone on the GuofengV1 and IWSLT2017 datasets.
As reported in Table \ref{tab:news}, \dragon consistently outperforms all baselines in the unseen literary and speech domains.
These findings suggest that the optimized context selection and utilization strategies are domain-agnostic rather than overfitting to the training data.
This enables \dragon to be deployed directly in diverse domains without additional tuning, underscoring its exceptional robustness and cross-domain generalization potential.

\begin{table}[t]
    \centering
    \scalebox{0.65}{
    \begin{tabular}{lcccccc}
        \toprule
        \multirow{2}{*}{System} & \multicolumn{3}{c}{Guofeng (Literary)} & \multicolumn{3}{c}{IWSLT (Speech)} \\
        \cmidrule(lr){2-4} \cmidrule(lr){5-7}
         & sCMT & dCMT & LLM & sCMT & dCMT & LLM \\
        \midrule
        Sentence        & 73.2 & 69.8 & 52.0 & 83.3 & 77.9 & 68.6 \\
        Segment         & 74.6 & 71.6 & 53.3 & 83.7 & 78.7 & 69.0 \\
        Doc2Doc         & 74.9 & 71.5 & 60.8 & 83.8 & 79.0 & 72.5 \\
        \textsc{DelTA}  & 74.2 & 70.5 & 63.0 & 83.4 & 78.4 & 71.0 \\
        \dragon         & \textbf{76.2} & \textbf{73.1} & \textbf{67.3} & \textbf{84.7} & \textbf{79.8} & \textbf{76.0} \\
        \bottomrule
    \end{tabular}}
    \caption{Test results on the OOD test sets.}
    \label{tab:news}
\end{table}

\paragraph{Stability on Long Document Translation}
To evaluate \dragon's adaptability to ultra-long texts, we use the En $\Rightarrow$ Xx Qwen3-8B to translate the first 12 chapters of the Chinese classical novel \textit{Journey to the West} into Portuguese. 
Considering the length, we set the retrieval sizes for the Essence and Exemplar components, $K_{\text{sum}}$ and $K_{\text{x}}$, to 8 and 6, respectively.
As shown in Figure \ref{fig:long_document}, Doc2Doc's accumulated history rapidly exceeds the model's context limit by Lines 156--160, causing translation failure. 
In contrast, \dragon's structured memory and selective retrieval significantly improve context utilization efficiency. 
Consequently, it seamlessly completes the entire document without interruption and consistently yields higher-quality translations than all baselines, demonstrating exceptional robustness in ultra-long scenarios.

\section*{Conclusion}

In this paper, we introduce \dragon, a human-like long-document translation agent with reasoning-driven adaptive context selection.
By leveraging a ``3E'' multi-granularity memory model, the agent captures and retrieves key contextual information in real time during sequential segment processing.
To mitigate interference from redundancy, \dragon employs an observe-and-act reasoning process to dynamically filter context candidates.
These strategies are further refined via RL based on sampled reasoning trajectories.
Empirical results demonstrate that \dragon achieves superior document translation quality across diverse language pairs and backbone models, and exhibits strong generalization ability to unseen languages and domains.

\section*{Limitations}

First, we currently adopt a fixed segmentation strategy (ten sentences per segment).
While effective, this static heuristic lacks the flexibility to align with natural discourse boundaries, which may limit the model's ability to maintain optimal coherence.
Future work could incorporate dynamic segmentation mechanisms to better capture discourse structures.
Second, the proposed agentic workflow, driven by deep reasoning, introduces additional computational overhead compared to standard one-pass generation.
We plan to address this efficiency bottleneck in future iterations through optimization techniques such as distillation or speculative decoding.
Finally, while we utilize COMET as the reward model for reinforcement learning, it may be an imperfect proxy for human preference in document-level translation tasks.
We look forward to adopting more robust and nuanced evaluation metrics as they become available to better align the agent's optimization with human judgments.

\bibliography{custom}

\appendix

\section{Appendix}

\subsection{Detailed Entity Attribute Fields}\label{sec:app_filed}

All entity categories and their corresponding attribute fields in the memory module are shown in Table \ref{tab:field}.
The extracted entities are classified into six different categories: \textit{Character}, \textit{Organization}, \textit{Location}, \textit{Event}, \textit{Object}, and \textit{Other}.
Each category has its unique attribute fields, reflecting different information focuses.

\subsection{Pseudo Codes}\label{app:code}

The pseudo code of the context-aware deep reasoning translation is shown in Algorithm \ref{alg:translation}.
During translation, the agent first roughly retrieves candidate information from the memory module (Lines 7-11), and then it conducts a deep reasoning selection to filter out the noisy items and keep the relevant ones as guidance for translating the current segment (Lines 12-19).
After the translation is done, the memory module is updated using the acquired results (Lines 20-24).

The pseudo code of the reasoning strategy optimization is shown in Algorithm \ref{alg:optimization}.
We first conduct parallel sampling on each reasoning step and the translation process to construct preference data (Lines 1-17), and then optimize the agent on it via reinforcement learning (Lines 18-21).

The sentence-level aligned translation algorithm is shown in Algorithm \ref{alg:recursive_translation}.
We insert boundary markers ``<s>'' and ``</s>'' around each sentence, and add sequential numbers ``\#i'' at the beginning (Lines 3-4).
The LLMs are told to keep these markers during translation (the corresponding prompt is shown in Figure \ref{fig:prompt_translation}). 
Then, we ask the agent to generate a target segment and check whether it is strictly aligned with the source at the sentence level (Lines 5-6).
If so, the results are output; otherwise, the source segment will be divided into two halves.
This procedure is executed recursively until alignment is achieved, or there's only one sentence left in the segment part (Lines 7-15).

\subsection{Case Study}\label{app:case_study}

We present a qualitative case study comparing the translation performance of \dragon against the sentence-level baseline utilizing Qwen3-8B as the base model.
Table \ref{tab:case_baseline} displays the source segment, the reference translation, and the baseline output.
Table \ref{tab:case_ours} details \dragon's translation workflow, illustrating its reasoning-driven context selection procedure and the final generation.

The introduction of Summary 4, which establishes the immediate scene of soldiers being awakened and equipped for action, significantly enhanced the contextual continuity of the translation. By recognizing that the current passage extends this preparation scene, the model maintained the narrative thread of ``coordination and readiness'' and avoided the disjointed framing seen in the Base version, which treated each line in isolation and consequently invented spurious military ranks (``Lieutenant Xiao'', ``Colonel Kolen'', ``Admiral Nimitz'').

Similarly, the incorporation of Sentence Pairs 3 and 4 directly supplies the canonical translations for the core characters and their actual ranks. Pair 4 in particular pins down ``\begin{CJK}{UTF8}{gbsn}科伦\end{CJK}'' as a \textit{lieutenant} (not a Colonel) and ``\begin{CJK}{UTF8}{gbsn}博拉丁·肖上尉\end{CJK}'' as \textit{Captain Borlatin Xiao}. With these in-context exemplars, the model produced ``Korren'' consistently across the entire window---in stark contrast to the baseline, which drifted through ``Cole'', ``Kolen'', and ``Korm'' within five sentences, and even mis-spelled the captain's name as ``Borlatan''.

Furthermore, the entity-level selection of records 5 and 9 demonstrates active filtering of redundant information: ten entities are pruned to two (i.e., the captain and the supporting armored squad). The model further detects that record 10 is redundant with record 5 and discards it, exhibiting an explicit anti-overfetching behavior. This focused entity grounding is what enables \dragon to render ``Captain Borlatin Xiao'' with the correct spelling and to anchor ``Nemic'' as a co-protagonist rather than fabricate the inflated ``Admiral Nimitz'' that the baseline produces.

\subsection{Detailed Dataset Information}\label{app:dataset}

The detailed information of the constructed training and test sets is shown in Table \ref{tab:trainingset} and \ref{tab:testset}, respectively.
To align with the application scenario of our long-document translation agent, we utilize the News Commentary V18.1 dataset (News domain) and first filter out documents with fewer than 50 lines.
Subsequently, we randomly sample 500 documents for each of the language pairs (English $\Leftrightarrow$ Chinese, German, and French) to construct the training set.
From the remaining data, we randomly sample 25 documents per language pair to form the test set.

For the unseen language pairs employed in \S\ref{sec:analysis}, we apply the same filtering strategy and randomly sample 25 documents for testing. The WMT24++ dataset encompasses four domains: News, Social, Speech, and Literary.
We filter the dataset to retain documents with $\geq$ 15 lines and then randomly sample 16 documents per language as test data.

For the IWSLT2017 (TED talks, speech domain) and GuofengV1 (webnovel, literary domain) datasets, we utilize their standard test sets without any additional filtering or sampling.

Finally, to evaluate ultra-long literary document translation, we use \textit{Journey to the West}, one of the Four Great Classical Novels of Chinese literature, which falls under the ultra-long Literary domain.
We utilize a Chinese-Portuguese parallel corpus of this novel and extract the first twelve chapters as our test data.

\subsection{Prompt Design}\label{app:prompt}

\begin{figure*}[h]
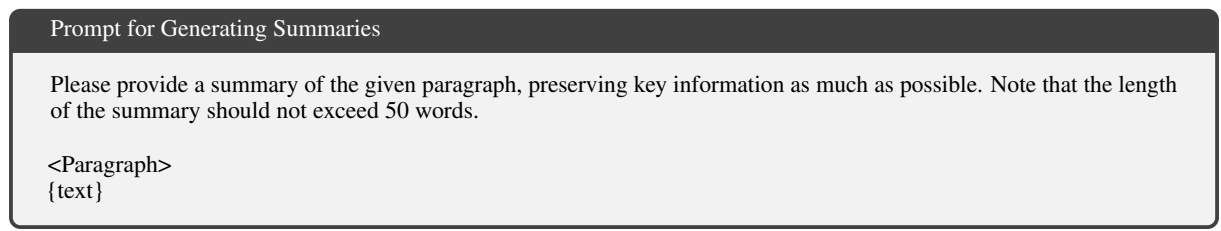

\centering
\small
\begin{tcolorbox}[title=Prompt for Generating Summaries]
Please provide a summary of the given paragraph, preserving key information as much as possible. Note that the length of the summary should not exceed 50 words.\\

<Paragraph>\\
\{text\}
\end{tcolorbox}
\caption{The prompt template for generating summaries.}
\label{fig:summary}
\end{figure*}

\begin{figure*}[h]
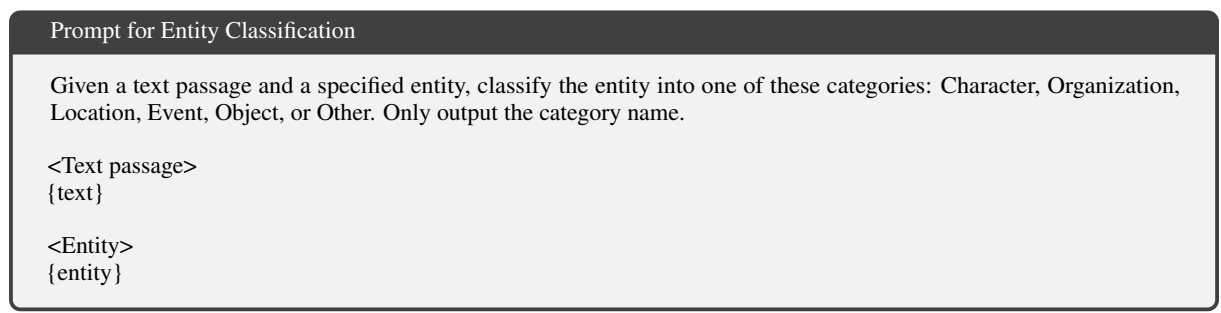

\centering
\small
\begin{tcolorbox}[title=Prompt for Entity Classification]
Given a text passage and a specified entity, classify the entity into one of these categories: Character, Organization, Location, Event, Object, or Other. Only output the category name.\\

<Text passage>\\
\{text\}\\

<Entity>\\
\{entity\}
\end{tcolorbox}
\caption{The prompt template for entity classification.}
\label{fig:entity_class}
\end{figure*}

\begin{figure*}[h]
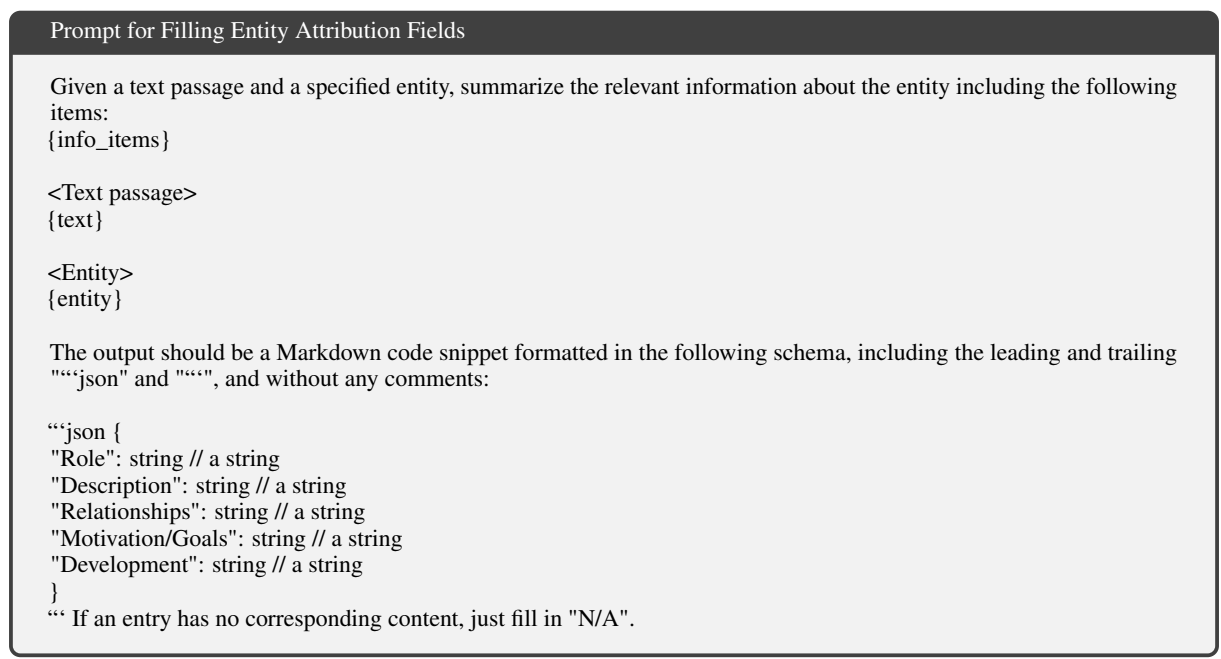

\centering
\small
\begin{tcolorbox}[title=Prompt for Filling Entity Attribution Fields]
Given a text passage and a specified entity, summarize the relevant information about the entity including the following items:\\
\{info\_items\}\\

<Text passage>\\
\{text\}\\

<Entity>\\
\{entity\}\\

The output should be a Markdown code snippet formatted in the following schema, including the leading and trailing "```json" and "```", and without any comments:\\

```json
\{\\
        "Role": string  // a string\\
        "Description": string  // a string\\
        "Relationships": string  // a string\\
        "Motivation/Goals": string  // a string\\
        "Development": string  // a string\\
\}\\
```
If an entry has no corresponding content, just fill in "N/A".
\end{tcolorbox}
\caption{The prompt template for filling entity attribution fields.}
\label{fig:entity_fill}
\end{figure*}

For the Essence memory component, the prompt for generating summaries is shown in Figure \ref{fig:summary}.
For the Entity memory component, the prompts of entity classification, attribute fields filling, and attribute fields update are shown in Figure \ref{fig:entity_class}, Figure \ref{fig:entity_fill}, and Figure \ref{fig:entity_update}, respectively.
The prompt for the final alignment-enforced document translation process of \dragon is shown in Figure \ref{fig:prompt_translation}.
The prompt of the LLM evaluator is illustrated in Figure \ref{fig:llm_judge}.

\subsection{Detailed Results}\label{app:results}

The detailed results of \dragon and the baselines in En $\Rightarrow$ Xx and Xx $\Rightarrow$ En tested on the News Commentary V18.1 data are shown in Table \ref{tab:detailed_results_news_commentary_en_xx} and Table \ref{tab:detailed_results_news_commentary_xx_en}, respectively.
Similarly, the corresponding results evaluated on the WMT24++ test set are presented in Table \ref{tab:detailed_results_wmt24pp_en_xx} and Table \ref{tab:detailed_results_wmt24pp_xx_en}.
Across nearly all evaluation metrics and backbone models, \dragon consistently achieves the best performance, demonstrating its effectiveness across different evaluation dimensions as well as its robustness across languages and backbone models.

\begin{figure*}[h]
\centering
\small
\begin{tcolorbox}[title=Prompt for Updating Entity Attribution Fields]
Given a text passage and a specified entity, update the existing information about this entity including the following items:\\
\{info\_items\}\\

<Text passage>\\
\{text\}\\

<Entity>\\
\{entity\}\\

<Existing Information>\\
\{exist\_info\}\\

```json
\{\\
        "Role": string  // a string\\
        "Description": string  // a string\\
        "Relationships": string  // a string\\
        "Motivation/Goals": string  // a string\\
        "Development": string  // a string\\
\}\\
```
If an entry has no corresponding content, just fill in "N/A".
\end{tcolorbox}
\caption{The prompt template for updating entity attribution fields.}
\label{fig:entity_update}
\end{figure*}

\begin{figure*}[h]
\centering
\small
\begin{tcolorbox}[title=Prompt for Document Translation by \dragon]
Given some auxiliary information, translate the current page of source text from \{src\_lang\} to \{tgt\_lang\}.

<Summaries of previous pages>\\
\{summaries\}\\

<Original texts of previous pages>\\
\{exemplars\}\\

<Entity Records>\\
\{Entities\}\\

Now please translate the given English text into Chinese. Make sure to obay the TRANSLATION TASK RULES.\\

<TRANSLATION TASK RULES>\\
1. Each sentence in the text is marked with ``\#i'' to indicate its order.\\
2. The beginning and end of an independent sentences are marked by ``<s>'' and ``</s>'', respectively.\\
3. Output MUST:\\
- Preserve ALL sequence, beginning and end marks (``\#i'', ``<s>'' and ``</s>'')\\
- Maintain EXACT 1:1 sentence correspondence\\
- NEVER merge/split/reorder/omit sentences\\

<\{src\_lang\} source text>\\
\{src\_content\}
\end{tcolorbox}
\caption{The prompt template for document translation by \dragon.}
\label{fig:prompt_translation}
\end{figure*}

\begin{figure*}[h]
\centering
\small
\begin{tcolorbox}[title=Prompt for LLM Evaluator]
You are an expert linguist and translation quality evaluator. Your task is to evaluate the quality of a document-level translation from \{src\_language\} to \{tgt\_language\} based solely on the Source Document, the Reference Document (Gold Standard), and the Hypothesis Document (Model Output).

Please assess the [Hypothesis] text as a whole against the [Source] and [Reference]. Provide a holistic score from 0 to 100 for the following three specific dimensions, where 0 represents a complete failure and 100 represents a perfect, native-level professional translation.

[Source]:
\{src\_doc\}

[Reference]:
\{ref\_doc\}

[Hypothesis]:
\{hyp\_doc\}

[Evaluation Dimensions]:

1. **General Quality**:\\
   - Focuses on accuracy (faithfulness to the source meaning) and fluency (grammatical correctness and natural flow).\\
   - A high score means the translation is precise, preserves the original meaning without omission or hallucination, and reads naturally in the target language.\\
2. **Cohesion**:\\
   - Focuses on the explicit linking words and grammatical connections between sentences and clauses (e.g., correct use of pronouns, conjunctions, substitution, and ellipsis).\\
   - A high score means the text is syntactically well-connected, and references (anaphora/cataphora) are clear and unambiguous throughout the document.\\
3. **Coherence**:\\
   - Focuses on the logical arrangement and semantic relationships of ideas. It assesses whether the text "makes sense" as a whole narrative or argument.\\
   - A high score means the discourse flows logically, follows the thought patterns/conventions of the target culture, and is easy for a reader to understand without referring to the source.\\
4. **Style Consistency**:\\
   - Focuses on the maintenance of tone, register (formal/informal), and voice throughout the document.\\
   - A high score means the translation maintains a unified style that matches the source text's intent (e.g., not switching between academic and slang phrasing).\\
5. **Terminology Consistency**:\\
   - Focuses on the consistent translation of specific terms, entities, and keywords across the entire document.\\
   - A high score means the same concept is translated using the same term throughout, avoiding confusion caused by using multiple synonyms for the same specific entity.

[Output Requirement]:
For each dimension, provide a score (0-100) and a brief justification based on the whole document.\\
Your response must strictly follow this format:\\
\#\#\# Evaluation Report\\
**1. General Quality**\\
Score: [0-100]\\
Rationale: ...\\
**2. Cohesion**\\
Score: [0-100]\\
Rationale: ...\\
**3. Coherence**\\
Score: [0-100]\\
Rationale: ...\\
**4. Style Consistency**\\
Score: [0-100]\\
Rationale: ...\\
**5. Terminology Consistency**\\
Score: [0-100]\\
Rationale: ...\\
\end{tcolorbox}
\caption{The prompt template of the LLM evaluator.}
\label{fig:llm_judge}
\end{figure*}

\clearpage

\begin{table*}[t]
    \centering
    \small
    \begin{tabular}{ll}
        \toprule
        Category & Attribute Fields \\
        \midrule
        Character       & Role, Description, Relationships, Motivation/Goals, Development \\
        Organization    & Type, Purpose, Members, Location, Significance \\
        Location        & Type, Description, Inhabitants, Events, Symbolism \\
        Event           & Title, Description, Participants, Location, Consequences, Timeline \\
        Object          & Type, Appearance, Purpose, Owner/Creator, Significance \\
        Other           & Label, Type, Description, Significance, Interaction, Impact \\
        \bottomrule
    \end{tabular}
    \caption{All entity categories and their corresponding attribute fields.}
    \label{tab:field}
\end{table*}

\begin{table*}[t]
    \centering
    \small
    \begin{tabular}{lcc}
        \toprule
        Model & Language & Size \\
        \midrule
        \multirow{2}{*}{Qwen2.5-7B-Instruct}  & En $\Rightarrow$ Xx & 86,429 \\
                                              & Xx $\Rightarrow$ En & 78,074 \\
        \hdashline
        \multirow{2}{*}{Qwen3-8B}             & En $\Rightarrow$ Xx & 82,324 \\
                                              & Xx $\Rightarrow$ En & 75,360 \\
        \hdashline
        \multirow{2}{*}{Qwen3-14B}            & En $\Rightarrow$ Xx & 78,832 \\
                                              & Xx $\Rightarrow$ En & 75,052 \\
        \hdashline
        \multirow{2}{*}{Llama3.1-8B-Instruct} & En $\Rightarrow$ Xx & 74,192 \\
                                              & Xx $\Rightarrow$ En & 79,210 \\
        \bottomrule
    \end{tabular}
    \caption{Statistics of the constructed training set for each model.}
    \label{tab:trainingset}
\end{table*}

\begin{table*}[t]
    \centering
    \small
    \begin{tabular}{lclccc}
        \toprule
        Dataset & Domain & Language & \# Words & \# Docs & Words / Doc. \\
        \midrule
        \multirow{9}{*}{News Commentary V18.1}  & \multirow{9}{*}{News} & English $\Leftrightarrow$ Chinese     & 24,284 & 25 & 971 \\
                                                &  & English $\Leftrightarrow$ German      & 24,369 & 25 & 975 \\
                                                &  & English $\Leftrightarrow$ French      & 24,813 & 25 & 993 \\
                                                &  & English $\Leftrightarrow$ Czech       & 23,961 & 25 & 958 \\
                                                &  & English $\Leftrightarrow$ Spanish     & 24,537 & 25 & 982 \\
                                                &  & English $\Leftrightarrow$ Italian     & 27,237 & 25 & 1,090 \\
                                                &  & English $\Leftrightarrow$ Japanese    & 20,054 & 25 & 802 \\
                                                &  & English $\Leftrightarrow$ Portuguese  & 26,775 & 25 & 1,071 \\
                                                &  & English $\Leftrightarrow$ Russian     & 24,472 & 25 & 979 \\
        \hdashline
        \multirow{3}{*}{WMT24++}                & \multirow{3}{*}{News, Social, Speech, Literary} & English $\Leftrightarrow$ Chinese     & 10,186 & 16 & 637 \\
                                                &  & English $\Leftrightarrow$ German      & 10,186 & 16 & 637 \\
                                                &  & English $\Leftrightarrow$ French      & 10,186 & 16 & 637 \\
        \hdashline
        \multirow{3}{*}{IWSLT2017}             & \multirow{3}{*}{Speech (TED talks)} & English $\Leftrightarrow$ Chinese     & 23,863 & 12 & 1,989 \\
                                                &  & English $\Leftrightarrow$ German      & 18,162 & 10 & 1,816 \\
                                                &  & English $\Leftrightarrow$ French      & 23,917 & 12 & 1,993 \\
        \hdashline
        GuoFengV1                              & Literary (Webnovels) & Chinese $\Rightarrow$ English         & 17,335 & 12 & 1,445 \\
        \hdashline
        Journey to the West                     & Literary (Ultra-Long novel) & Chinese $\Rightarrow$ Portuguese      & 51,854 & 1 & 51,854 \\
        \bottomrule
    \end{tabular}
    \caption{Statistics of the adopted test sets.}
    \label{tab:testset}
\end{table*}

\begin{table*}[h]
    \centering
    \scalebox{0.85}{
    \adjustbox{center}{
        \begin{tabular}{ll cccc cccc cccc}
            \toprule
            \multirow{2}{*}{Model} & \multirow{2}{*}{System} & \multicolumn{4}{c}{En $\Rightarrow$ Zh} & \multicolumn{4}{c}{En $\Rightarrow$ De} & \multicolumn{4}{c}{En $\Rightarrow$ Fr} \\
            \cmidrule(lr){3-6} \cmidrule(lr){7-10} \cmidrule(lr){11-14}
             & & sCMT & dCMT & LLM & Avg & sCMT & dCMT & LLM & Avg & sCMT & dCMT & LLM & Avg \\
            \midrule
            \multirow{5}{*}{\shortstack[c]{Qwen2.5\\7B}}
             & Sentence        & 87.6 & 83.0 & 82.9 & 84.5 & 82.9 & 77.3 & 40.8 & 67.0 & 82.8 & 77.0 & 45.0 & 68.3 \\
             & Segment         & 88.0 & 83.6 & \textbf{88.9} & \textbf{86.8} & 82.7 & 77.1 & 43.0 & 67.6 & 83.8 & 78.1 & 60.2 & 74.0 \\
             & Doc2Doc         & 87.5 & 82.8 & 85.3 & 85.2 & 82.7 & 77.1 & 43.7 & 67.8 & 83.3 & 77.5 & 60.3 & 73.7 \\
             & \textsc{DelTA}  & 87.9 & 83.3 & 82.5 & 84.6 & 83.5 & 77.8 & 44.5 & 68.6 & 83.9 & 78.3 & 63.3 & 75.2 \\
             & \dragon         & \textbf{88.2} & \textbf{83.7} & 85.1 & 85.7 & \textbf{84.6} & \textbf{78.7} & \textbf{49.4} & \textbf{70.9} & \textbf{84.8} & \textbf{79.1} & \textbf{65.7} & \textbf{76.5} \\
            \midrule
            \multirow{5}{*}{\shortstack[c]{Qwen3\\8B}}
             & Sentence        & 88.5 & 84.0 & 92.4 & 88.3 & 86.5 & 80.9 & 67.8 & 78.4 & 85.5 & 80.0 & 76.5 & 80.7 \\
             & Segment         & 88.7 & 84.3 & 93.4 & 88.8 & 85.6 & 79.0 & 64.9 & 76.5 & 85.3 & 79.4 & 76.5 & 80.4 \\
             & Doc2Doc         & 88.7 & 84.2 & \textbf{93.6} & \textbf{88.8} & 85.7 & 79.3 & 66.0 & 77.0 & 85.0 & 78.7 & 78.1 & 80.6 \\
             & \textsc{DelTA}  & 88.3 & 83.8 & 91.1 & 87.7 & 86.3 & 80.6 & 66.3 & 77.7 & 85.5 & 79.9 & 78.9 & 81.4 \\
             & \dragon         & \textbf{88.9} & \textbf{84.5} & 93.1 & 88.8 & \textbf{86.9} & \textbf{81.2} & \textbf{69.6} & \textbf{79.2} & \textbf{86.1} & \textbf{80.5} & \textbf{82.5} & \textbf{83.1} \\
            \midrule
            \multirow{5}{*}{\shortstack[c]{Qwen3\\14B}}
             & Sentence        & 88.8 & 84.3 & 93.1 & 88.7 & 87.4 & 81.8 & 76.3 & 81.8 & 86.2 & 80.6 & 83.8 & 83.5 \\
             & Segment         & 88.9 & \textbf{84.6} & \textbf{94.5} & \textbf{89.3} & 87.2 & 81.5 & 72.9 & 80.5 & 85.9 & 80.5 & 83.3 & 83.2 \\
             & Doc2Doc         & 88.9 & 84.5 & 94.1 & 89.2 & 87.3 & 81.8 & 75.7 & 81.6 & 85.9 & 80.3 & 84.5 & 83.6 \\
             & \textsc{DelTA}  & 88.8 & 84.3 & 94.2 & 89.1 & 87.4 & 81.9 & 75.6 & 81.6 & 86.1 & 80.6 & 85.2 & 84.0 \\
             & \dragon         & \textbf{89.0} & 84.6 & 92.0 & 88.5 & \textbf{87.9} & \textbf{82.3} & \textbf{79.2} & \textbf{83.1} & \textbf{86.4} & \textbf{80.9} & \textbf{85.5} & \textbf{84.3} \\
            \midrule
            \multirow{5}{*}{\shortstack[c]{Llama3.1\\8B}}
             & Sentence        & 85.0 & 77.6 & 65.8 & 76.1 & 85.2 & 79.5 & 60.9 & 75.2 & 84.3 & 78.8 & 69.4 & 77.5 \\
             & Segment         & 84.1 & 77.2 & 61.3 & 74.2 & 84.8 & 79.3 & 58.2 & 74.1 & 84.0 & 78.4 & 63.9 & 75.4 \\
             & Doc2Doc         & 85.0 & 79.5 & 66.5 & 77.0 & 85.1 & 79.5 & 62.1 & 75.6 & 83.7 & 78.2 & 65.8 & 75.9 \\
             & \textsc{DelTA}  & 85.8 & 79.8 & 75.0 & 80.2 & 85.9 & 80.3 & 66.4 & 77.5 & 84.8 & 79.2 & 72.1 & 78.7 \\
             & \dragon         & \textbf{87.0} & \textbf{82.6} & \textbf{76.5} & \textbf{82.0} & \textbf{86.6} & \textbf{81.2} & \textbf{69.8} & \textbf{79.2} & \textbf{85.5} & \textbf{80.1} & \textbf{72.8} & \textbf{79.4} \\
            \bottomrule
        \end{tabular}
    }}
    \caption{Detailed results of \dragon compared to the baseline methods in En $\Rightarrow$ Xx directions on the \textbf{News Commentary V18.1} test set. The best results are marked in \textbf{bold font}.}
    \label{tab:detailed_results_news_commentary_en_xx}
\end{table*}

\begin{table*}[h]
    \centering
    \scalebox{0.85}{
    \adjustbox{center}{
        \begin{tabular}{ll cccc cccc cccc}
            \toprule
            \multirow{2}{*}{Model} & \multirow{2}{*}{System} & \multicolumn{4}{c}{Zh $\Rightarrow$ En} & \multicolumn{4}{c}{De $\Rightarrow$ En} & \multicolumn{4}{c}{Fr $\Rightarrow$ En} \\
            \cmidrule(lr){3-6} \cmidrule(lr){7-10} \cmidrule(lr){11-14}
             & & sCMT & dCMT & LLM & Avg & sCMT & dCMT & LLM & Avg & sCMT & dCMT & LLM & Avg \\
            \midrule
            \multirow{5}{*}{\shortstack[c]{Qwen2.5\\7B}}
             & Sentence        & 86.2 & 82.3 & 81.0 & 83.2 & 87.4 & 82.8 & 78.8 & 83.0 & 86.5 & 81.1 & 75.4 & 81.0 \\
             & Segment         & 86.3 & 81.9 & 83.5 & 83.9 & 86.3 & 77.7 & 71.2 & 78.4 & 87.0 & 81.2 & 84.0 & 84.0 \\
             & Doc2Doc         & 85.7 & 81.0 & 80.1 & 82.3 & 86.4 & 78.1 & 71.8 & 78.8 & 85.1 & 75.9 & 70.9 & 77.3 \\
             & \textsc{DelTA}  & 86.7 & 83.0 & 87.1 & 85.6 & 88.3 & 83.9 & 87.3 & 86.5 & 87.4 & 82.3 & 88.3 & 86.0 \\
             & \dragon         & \textbf{87.4} & \textbf{83.7} & \textbf{91.5} & \textbf{87.5} & \textbf{88.9} & \textbf{84.4} & \textbf{90.3} & \textbf{87.9} & \textbf{87.9} & \textbf{83.0} & \textbf{90.4} & \textbf{87.1} \\
            \midrule
            \multirow{5}{*}{\shortstack[c]{Qwen3\\8B}}
             & Sentence        & 86.8 & 83.0 & 86.4 & 85.4 & 88.5 & 84.1 & 88.8 & 87.1 & 87.4 & 82.3 & 88.1 & 86.0 \\
             & Segment         & 87.0 & 83.2 & 87.2 & 85.8 & 87.8 & 81.7 & 79.9 & 83.1 & 86.9 & 80.6 & 80.7 & 82.7 \\
             & Doc2Doc         & 87.2 & 83.5 & 88.0 & 86.2 & 88.3 & 82.7 & 85.0 & 85.3 & 87.2 & 81.5 & 85.5 & 84.7 \\
             & \textsc{DelTA}  & 87.1 & 83.3 & 88.6 & 86.3 & 88.7 & 84.3 & 89.3 & 87.4 & 87.5 & 82.5 & 89.3 & 86.4 \\
             & \dragon         & \textbf{87.3} & \textbf{83.7} & \textbf{90.2} & \textbf{87.1} & \textbf{89.0} & \textbf{84.5} & \textbf{90.8} & \textbf{88.1} & \textbf{87.8} & \textbf{82.8} & \textbf{89.7} & \textbf{86.8} \\
            \midrule
            \multirow{5}{*}{\shortstack[c]{Qwen3\\14B}}
             & Sentence        & 87.2 & 83.4 & 88.0 & 86.2 & 88.7 & 84.3 & 89.1 & 87.4 & 87.7 & 82.8 & 87.6 & 86.0 \\
             & Segment         & 87.3 & 83.7 & 89.7 & 86.9 & 88.9 & 84.5 & 89.5 & 87.6 & 87.6 & 82.7 & 88.1 & 86.1 \\
             & Doc2Doc         & 87.4 & 83.9 & 90.5 & 87.3 & 89.0 & 84.6 & 90.1 & 87.9 & 87.7 & 82.9 & 89.6 & 86.7 \\
             & \textsc{DelTA}  & 87.4 & 83.8 & 90.6 & 87.3 & 88.9 & 84.5 & 90.5 & 87.9 & 87.8 & 82.9 & 90.1 & 87.0 \\
             & \dragon         & \textbf{87.7} & \textbf{84.1} & \textbf{91.1} & \textbf{87.6} & \textbf{89.2} & \textbf{85.0} & \textbf{91.1} & \textbf{88.5} & \textbf{88.0} & \textbf{83.1} & \textbf{91.1} & \textbf{87.4} \\
            \midrule
            \multirow{5}{*}{\shortstack[c]{Llama3.1\\8B}}
             & Sentence        & 85.7 & 81.7 & 78.1 & 81.8 & 88.0 & 83.6 & 84.3 & 85.3 & 86.8 & 81.7 & 83.2 & 83.9 \\
             & Segment         & 85.8 & 82.0 & 79.0 & 82.2 & 88.1 & 83.6 & 84.9 & 85.5 & 86.9 & 81.6 & 82.1 & 83.5 \\
             & Doc2Doc         & 85.8 & 81.9 & 79.1 & 82.3 & 87.9 & 82.9 & 83.2 & 84.7 & 86.9 & 81.8 & 84.2 & 84.3 \\
             & \textsc{DelTA}  & 85.7 & 81.9 & 83.1 & 83.6 & 87.9 & 83.5 & 86.3 & 85.9 & 86.8 & 81.6 & 85.2 & 84.5 \\
             & \dragon         & \textbf{86.8} & \textbf{83.1} & \textbf{84.9} & \textbf{85.0} & \textbf{88.9} & \textbf{84.5} & \textbf{88.6} & \textbf{87.3} & \textbf{87.7} & \textbf{82.8} & \textbf{89.3} & \textbf{86.6} \\
            \bottomrule
        \end{tabular}
    }}
    \caption{Detailed results of \dragon compared to the baseline methods in Xx $\Rightarrow$ En directions on the \textbf{News Commentary V18.1} test set. The best results are marked in \textbf{bold font}.}
    \label{tab:detailed_results_news_commentary_xx_en}
\end{table*}

\begin{table*}[h]
    \centering
    \scalebox{0.85}{
    \adjustbox{center}{
        \begin{tabular}{ll cccc cccc cccc}
            \toprule
            \multirow{2}{*}{Model} & \multirow{2}{*}{System} & \multicolumn{4}{c}{En $\Rightarrow$ Zh} & \multicolumn{4}{c}{En $\Rightarrow$ De} & \multicolumn{4}{c}{En $\Rightarrow$ Fr} \\
            \cmidrule(lr){3-6} \cmidrule(lr){7-10} \cmidrule(lr){11-14}
             & & sCMT & dCMT & LLM & Avg & sCMT & dCMT & LLM & Avg & sCMT & dCMT & LLM & Avg \\
            \midrule
            \multirow{5}{*}{\shortstack[c]{Qwen2.5\\7B}}
             & Sentence        & 80.3 & 78.2 & 62.6 & 73.7 & 71.9 & 68.3 & 37.9 & 59.4 & 70.8 & 66.8 & 34.3 & 57.3 \\
             & Segment         & 83.6 & 81.7 & 79.0 & 81.4 & 77.1 & 74.0 & 48.6 & 66.6 & 78.2 & 75.3 & 60.4 & 71.3 \\
             & Doc2Doc         & 83.5 & 81.2 & \textbf{80.1} & 81.6 & 78.9 & 75.4 & 52.3 & 68.9 & 78.0 & 75.6 & 62.3 & 72.0 \\
             & \textsc{DelTA}  & 83.8 & 82.0 & 77.7 & 81.2 & 77.2 & 74.7 & 52.2 & 68.0 & 79.1 & 75.9 & \textbf{65.8} & \textbf{73.6} \\
             & \dragon         & \textbf{84.6} & \textbf{82.8} & 77.5 & \textbf{81.6} & \textbf{79.5} & \textbf{76.9} & \textbf{53.0} & \textbf{69.8} & \textbf{79.2} & \textbf{76.4} & 64.4 & 73.3 \\
            \midrule
            \multirow{5}{*}{\shortstack[c]{Qwen3\\8B}}
             & Sentence        & 84.4 & 82.6 & 80.7 & 82.6 & 80.9 & 78.0 & 66.1 & 75.0 & 80.2 & 77.5 & 70.7 & 76.1 \\
             & Segment         & 84.3 & 81.3 & 80.1 & 81.9 & 78.8 & 74.7 & 62.7 & 72.0 & 80.3 & 76.4 & 69.9 & 75.5 \\
             & Doc2Doc         & 84.9 & 82.9 & 84.8 & 84.2 & 80.7 & 77.6 & 68.9 & 75.7 & 80.3 & 76.6 & 69.5 & 75.5 \\
             & \textsc{DelTA}  & 84.6 & 83.1 & \textbf{85.2} & 84.3 & 80.6 & 78.2 & 67.4 & 75.4 & 80.3 & 77.3 & 72.7 & 76.7 \\
             & \dragon         & \textbf{85.7} & \textbf{83.8} & 84.8 & \textbf{84.8} & \textbf{81.7} & \textbf{78.9} & \textbf{69.5} & \textbf{76.7} & \textbf{81.0} & \textbf{78.4} & \textbf{76.0} & \textbf{78.5} \\
            \midrule
            \multirow{5}{*}{\shortstack[c]{Qwen3\\14B}}
             & Sentence        & 85.1 & 83.7 & 85.6 & 84.8 & 81.9 & 79.7 & 73.6 & 78.4 & 81.3 & 78.2 & 78.5 & 79.3 \\
             & Segment         & 85.0 & 82.6 & 84.6 & 84.0 & 82.3 & 79.4 & 72.8 & 78.1 & 81.5 & 77.5 & 79.0 & 79.3 \\
             & Doc2Doc         & 85.1 & 83.2 & 88.3 & 85.5 & 82.6 & 79.8 & \textbf{74.4} & 78.9 & 81.8 & 78.4 & 78.9 & 79.7 \\
             & \textsc{DelTA}  & 85.2 & \textbf{83.7} & 87.8 & 85.5 & 82.0 & 79.4 & 74.3 & 78.5 & 81.5 & \textbf{78.6} & \textbf{79.9} & \textbf{80.0} \\
             & \dragon         & \textbf{85.6} & 83.6 & \textbf{88.3} & \textbf{85.8} & \textbf{82.7} & \textbf{80.4} & 74.2 & \textbf{79.1} & \textbf{81.8} & 78.4 & 79.1 & 79.8 \\
            \midrule
            \multirow{5}{*}{\shortstack[c]{Llama3.1\\8B}}
             & Sentence        & 78.0 & 74.0 & 55.3 & 69.1 & 76.4 & 73.7 & 53.3 & 67.8 & 75.8 & 73.4 & 55.5 & 68.2 \\
             & Segment         & 76.9 & 70.4 & 54.3 & 67.2 & 79.1 & 76.2 & 56.6 & 70.6 & 77.8 & 75.1 & 56.3 & 69.7 \\
             & Doc2Doc         & 80.3 & 78.7 & 61.4 & 73.5 & 78.9 & 76.4 & 57.2 & 70.9 & 77.6 & 75.9 & 61.1 & 71.5 \\
             & \textsc{DelTA}  & 79.7 & 77.9 & \textbf{63.6} & 73.7 & 78.5 & 75.9 & 59.9 & 71.4 & 77.5 & 75.2 & 62.7 & 71.8 \\
             & \dragon         & \textbf{81.6} & \textbf{79.8} & 63.0 & \textbf{74.8} & \textbf{80.2} & \textbf{77.2} & \textbf{63.5} & \textbf{73.6} & \textbf{79.6} & \textbf{76.8} & \textbf{66.7} & \textbf{74.3} \\
            \bottomrule
        \end{tabular}
    }}
    \caption{Detailed results of \dragon compared to the baseline methods in En $\Rightarrow$ Xx directions on the \textbf{WMT24++} test set. The best results are marked in \textbf{bold font}.}
    \label{tab:detailed_results_wmt24pp_en_xx}
\end{table*}

\begin{table*}[h]
    \centering
    \scalebox{0.85}{
    \adjustbox{center}{
        \begin{tabular}{ll cccc cccc cccc}
            \toprule
            \multirow{2}{*}{Model} & \multirow{2}{*}{System} & \multicolumn{4}{c}{Zh $\Rightarrow$ En} & \multicolumn{4}{c}{De $\Rightarrow$ En} & \multicolumn{4}{c}{Fr $\Rightarrow$ En} \\
            \cmidrule(lr){3-6} \cmidrule(lr){7-10} \cmidrule(lr){11-14}
             & & sCMT & dCMT & LLM & Avg & sCMT & dCMT & LLM & Avg & sCMT & dCMT & LLM & Avg \\
            \midrule
            \multirow{5}{*}{\shortstack[c]{Qwen2.5\\7B}}
             & Sentence        & 78.3 & 75.3 & 56.4 & 70.0 & 78.0 & 74.9 & 59.3 & 70.7 & 77.2 & 73.5 & 56.4 & 69.0 \\
             & Segment         & 82.6 & 79.8 & 75.2 & 79.2 & 82.2 & 75.6 & 69.1 & 75.6 & 83.5 & 79.2 & 77.1 & 79.9 \\
             & Doc2Doc         & 82.8 & 80.2 & 76.9 & 80.0 & 82.7 & 77.2 & 71.4 & 77.1 & 83.4 & 78.9 & 76.1 & 79.5 \\
             & \textsc{DelTA}  & 82.9 & 80.4 & 76.9 & 80.1 & 84.7 & 82.2 & 82.5 & 83.1 & 84.1 & 81.7 & 81.7 & 82.5 \\
             & \dragon         & \textbf{83.3} & \textbf{81.0} & \textbf{79.8} & \textbf{81.4} & \textbf{85.5} & \textbf{83.3} & \textbf{83.0} & \textbf{84.0} & \textbf{84.4} & \textbf{82.0} & \textbf{83.3} & \textbf{83.2} \\
            \midrule
            \multirow{5}{*}{\shortstack[c]{Qwen3\\8B}}
             & Sentence        & 82.8 & 80.0 & 74.3 & 79.0 & 84.2 & 81.4 & 77.4 & 81.0 & 83.4 & 81.3 & 75.6 & 80.1 \\
             & Segment         & 83.7 & 81.2 & 77.0 & 80.6 & 83.7 & 78.3 & 72.2 & 78.1 & 83.1 & 78.2 & 73.6 & 78.3 \\
             & Doc2Doc         & 83.6 & 80.9 & 80.6 & 81.7 & 84.4 & 81.0 & 78.7 & 81.4 & 83.6 & 79.4 & 76.4 & 79.8 \\
             & \textsc{DelTA}  & 83.4 & 80.9 & 78.6 & 81.0 & 84.8 & 82.5 & 82.0 & 83.1 & 84.4 & 82.3 & 82.8 & 83.1 \\
             & \dragon         & \textbf{84.2} & \textbf{81.8} & \textbf{80.8} & \textbf{82.2} & \textbf{85.5} & \textbf{83.1} & \textbf{85.4} & \textbf{84.7} & \textbf{84.8} & \textbf{82.5} & \textbf{84.5} & \textbf{83.9} \\
            \midrule
            \multirow{5}{*}{\shortstack[c]{Qwen3\\14B}}
             & Sentence        & 83.4 & 80.8 & 76.5 & 80.2 & 85.1 & 82.6 & 83.2 & 83.6 & 84.5 & 82.1 & 81.6 & 82.7 \\
             & Segment         & 84.1 & 81.8 & 79.8 & 81.9 & 85.6 & 83.3 & 83.9 & 84.3 & 84.9 & 82.5 & 84.1 & 83.8 \\
             & Doc2Doc         & 84.0 & 81.9 & \textbf{83.9} & \textbf{83.3} & 85.6 & 83.3 & 85.3 & 84.7 & 85.1 & 82.7 & 86.2 & 84.6 \\
             & \textsc{DelTA}  & 83.9 & 81.8 & 83.7 & 83.1 & 85.5 & 82.9 & 84.3 & 84.2 & 84.9 & 82.5 & 85.1 & 84.2 \\
             & \dragon         & \textbf{84.2} & \textbf{81.9} & 82.3 & 82.8 & \textbf{85.7} & \textbf{83.4} & \textbf{87.4} & \textbf{85.5} & \textbf{85.2} & \textbf{82.9} & \textbf{86.3} & \textbf{84.8} \\
            \midrule
            \multirow{5}{*}{\shortstack[c]{Llama3.1\\8B}}
             & Sentence        & 79.8 & 76.9 & 65.9 & 74.2 & 81.3 & 78.3 & 73.0 & 77.5 & 80.2 & 77.8 & 69.9 & 76.0 \\
             & Segment         & 82.5 & 79.8 & 72.3 & 78.2 & 84.3 & 81.5 & 77.8 & 81.2 & 83.9 & 81.2 & 73.5 & 79.5 \\
             & Doc2Doc         & 82.4 & 80.1 & 75.2 & 79.2 & 84.7 & 81.6 & 78.3 & 81.5 & 83.7 & 81.2 & 76.6 & 80.5 \\
             & \textsc{DelTA}  & 78.1 & 74.7 & 70.6 & 74.5 & 80.7 & 77.9 & 78.5 & 79.0 & 80.2 & 77.5 & 76.7 & 78.1 \\
             & \dragon         & \textbf{83.0} & \textbf{80.3} & \textbf{78.5} & \textbf{80.6} & \textbf{85.1} & \textbf{82.8} & \textbf{84.9} & \textbf{84.3} & \textbf{84.3} & \textbf{81.9} & \textbf{83.6} & \textbf{83.3} \\
            \bottomrule
        \end{tabular}
    }}
    \caption{Detailed results of \dragon compared to the baseline methods in Xx $\Rightarrow$ En directions on the \textbf{WMT24++} test set. The best results are marked in \textbf{bold font}.}
    \label{tab:detailed_results_wmt24pp_xx_en}
\end{table*}

\begin{algorithm*}[t]
\caption{Context-Aware Deep Reasoning Translation with Memory (\dragon)}
\label{alg:translation}
\SetKwInOut{Input}{Input}
\SetKwInOut{Output}{Output}
\SetKwFunction{LLM}{LLM}
\SetKwFunction{Embed}{Embed}
\SetKwFunction{Sim}{Sim}
\SetKwFunction{TopK}{TopK}
\SetKwFunction{ExtractEntities}{ExtractEntities}
\SetKwFunction{Summarize}{Summarize}

\Input{Source Document $S$, Segment size $l$, LLM $\mathcal{M}$}
\Output{Target Document $T$}

\BlankLine
\tcc{Initialize Memory Components}
$\mathcal{E}_\text{s}^{(0)} \leftarrow \emptyset$ \tcp*[r]{Essence}
$\mathcal{E}_\text{x}^{(0)} \leftarrow \emptyset$ \tcp*[r]{Exemplars}
$\mathcal{E}_\text{n}^{(0)} \leftarrow \emptyset$ \tcp*[r]{Entities}

\BlankLine
\tcc{Segmentation}
Split $S$ into $L$ segments $\{\bm{s}_1, \bm{s}_2, \dots, \bm{s}_L\}$, where $|\bm{s}_\tau| = l$\;
$T \leftarrow \emptyset$\;

\BlankLine
\For{$t \leftarrow 1$ \KwTo $N$}{
    \tcc{Step 1: Memory Retrieval}
    $\mathbf{v}_{\tau} \leftarrow \Embed(\bm{s}_\tau)$\;
    $\tilde{\mathcal{E}}_\text{s} \leftarrow \TopK_{\bm{a} \in \mathcal{E}_\text{s}^{(\tau-1)}} \Sim(\Embed(\bm{a}), \mathbf{v}_{\tau}, K_{\text{s}})$\;
    $\tilde{\mathcal{E}}_\text{x} \leftarrow \TopK_{(\bm{u}, \cdot) \in \mathcal{E}_\text{c}^{(\tau-1)}} \Sim(\Embed(\bm{u}), \mathbf{v}_{\tau}, K_{\text{x}})$\;
    $\tilde{\mathcal{E}}_\text{n} \leftarrow \{ (e^\text{src}, e^\text{tgt}, \LLM(\text{Desc}, \mathcal{I}, \bm{s}_\tau)) \mid (e^\text{src}, e^\text{tgt}, \mathcal{I}) \in \mathcal{E}_\text{n}^{(\tau-1)}, e^\text{src} \in \bm{s}_\tau \}$\;
    
    Define Candidate Set $\mathbf{E} \leftarrow [\tilde{\mathcal{E}_\text{s}}, \tilde{\mathcal{E}_\text{x}}, \tilde{\mathcal{E}_\text{n}}]$\;
    
    \BlankLine
    \tcc{Step 2: Deep Reasoning Translation}
    $\mathcal{H}_0 \leftarrow \emptyset$ \tcp*[r]{Reasoning history}
    
    \For{$k \leftarrow 1$ \KwTo $3$}{
        \tcc{Observe: History, Specific Context}
        $\mathcal{O}_k \leftarrow \{ \mathcal{H}_{k-1}, \mathbf{E}[k] \}$\;
        
        \tcc{Action Sub-steps: Analyze, Choose, Translate}
        $\bm{r}_k \leftarrow \LLM(\text{Analyze}, \mathcal{O}_k)$ \tcp*[r]{Assess usefulness}
        $\mathcal{C}_k \leftarrow \LLM(\text{Choose}, \bm{r}_k)$ \tcp*[r]{Select optimal context}
        
        $\mathcal{H}_k \leftarrow \mathcal{H}_{k-1} \cup \{ (\mathcal{O}_k, \langle \bm{r}_k, \mathcal{C}_k) \rangle \}$\;
    }
    $\bm{t}_\tau \leftarrow \LLM(\text{Translate}, \bm{s}_\tau, \{\mathcal{C}_k\}_{k=1}^3)$\;
    Append $\bm{t}_\tau$ to $T$\;

    \BlankLine
    \tcc{Step 3: Memory Update}
    $\bm{a}_{\text{new}} \leftarrow \Summarize(\bm{s}_\tau)$\;
    $\mathcal{E}_\text{s}^{(\tau)} \leftarrow \mathcal{E}_\text{s}^{(\tau-1)} \cup \{ \bm{a}_{\text{new}} \}$\;
    
    $\mathcal{E}_\text{x}^{(\tau)} \leftarrow \mathcal{E}_\text{x}^{(\tau-1)} \cup \{(\bm{u}, \bm{v}) \mid \bm{u} \in \bm{s}_\tau, \bm{v} \in \bm{t}_\tau\}$\;
    
    $\mathcal{E}_\text{n}^{\text{new}} \leftarrow \ExtractEntities(\bm{s}_\tau, \bm{t}_\tau)$\;
    $\mathcal{E}_\text{n}^{(\tau)} \leftarrow \text{UpdateRecords}(\mathcal{E}_\text{n}^{(\tau-1)}, \mathcal{E}_\text{n}^{\text{new}})$\;
}
\Return{$T$}
\end{algorithm*}

\begin{algorithm*}[t]
\caption{Reasoning Strategy Optimization}
\label{alg:optimization}
\SetKwInOut{Input}{Input}
\SetKwInOut{Output}{Output}
\SetKwFunction{Sample}{Sample}
\SetKwFunction{Comet}{COMET}
\SetKwFunction{TrainSFT}{TrainSFT}
\SetKwFunction{TrainDPO}{TrainDPO}
\SetKwFunction{ArgMax}{arg\,max}
\SetKwFunction{ArgMin}{arg\,min}

\Input{Original Training Set $\mathcal{D}_\text{raw}$, Pretrained LLM $\bm{\pi}_{\theta}$}
\Output{Optimized LLM $\bm{\pi}_{\theta^*}$}

\BlankLine
\tcc{Phase 1: Observe-and-Act Data Construction}
Initialize datasets $\mathcal{D}_{\text{sel}} \leftarrow \emptyset$, $\mathcal{D}_{\text{util}} \leftarrow \emptyset$\;

\For{each pair $(\bm{s}, \bm{t}^*)$ in $\mathcal{D}_\textup{raw}$}{
    \For{reasoning step $k \leftarrow 1$ \KwTo $3$}{
        \tcc{Sample Actions Parallelly for $M$ times}
        $\mathcal{G}_k \leftarrow \Sample(\bm{\bm{\pi}}_{\theta}, \mathcal{O}_k, M)$\;
        
        \For{each action $\mathcal{A}_k^{i}=\langle \bm{r}_k^{i}, \mathcal{C}_k^{i} \rangle \in \mathcal{G}_k$}{
            \tcc{Sample Translations for $N$ times \& Evaluate}
            $\mathcal{T}_k^{i} \leftarrow \Sample(\text{Translate}, \bm{s}, \mathcal{C}_k^{i},N)$\;
            $U(\mathcal{A}_k^{i}) \leftarrow \frac{1}{N} \sum_{\hat{\bm{t}} \in \mathcal{T}_k^{i}} \Comet(\bm{s}, \hat{\bm{t}}, \bm{t}^*)$\;
            
            \tcc{Construct Translation Preference Data}
            $\bm{t}^+ \leftarrow \ArgMax_{\bm{t} \in \mathcal{T}_k^{i}} \Comet(\bm{s}, \hat{\bm{t}}, \bm{t}^*)$\;
            $\bm{t}^- \leftarrow \ArgMin_{\bm{t} \in \mathcal{T}_k^{i}} \Comet(\bm{s}, \hat{\bm{t}}, \bm{t}^*)$\;
            $\mathcal{D}_{\text{util}} \leftarrow \mathcal{D}_{\text{util}} \cup \{ (\langle\bm{s}, \mathcal{C}_k^{i}\rangle, \bm{t}^+, \bm{t}^-) \}$\;
        }
        
        \tcc{Construct Context-Strategic Preference Data}
        $\mathcal{A}_k^+ \leftarrow \mathcal{G}_k[\ArgMax(U(\mathcal{A}))]$ \tcp*[r]{Preferred Action}
        $\mathcal{A}_k^- \leftarrow \mathcal{G}_k[\ArgMin(U(\mathcal{A}))]$ \tcp*[r]{Dispreferred Action}
        $\mathcal{D}_{\text{sel}} \leftarrow \mathcal{D}_{\text{sel}} \cup \{ (\mathcal{O}_k, \mathcal{A}_k^+, \mathcal{A}_k^-) \}$\;
        
        \tcc{Update State for Next Step}
        Identify $\bm{t}^+$ associated with $\mathcal{A}_k^+$\;
        Construct $\mathcal{O}_{k+1}$ using $\mathcal{A}_i^+$\;
        Update Memory using $\bm{t}^+$\;
    }
}

\BlankLine
\tcc{Phase 2: Training (SFT + DPO)}
Merge datasets: $\mathcal{D} \leftarrow \mathcal{D}_{\text{sel}} \cup \mathcal{D}_{\text{util}}$\;

\tcp{Stage 1: Cold-start SFT}
Extract preferred samples: $\mathcal{D}_{\text{SFT}} \leftarrow \{ (\bm{x}, \bm{y}^+) \mid (\bm{x}, \bm{y}^+, \bm{y}^-) \in \mathcal{D} \}$\;
$\bm{\pi}_{\text{SFT}} \leftarrow \TrainSFT(\bm{\pi}_{\theta}, \mathcal{D}_{\text{SFT}})$\;

\tcp{Stage 2: DPO with LoRA}
$\bm{\pi}_{\theta^*} \leftarrow \TrainDPO(\bm{\pi}_{\text{SFT}}, \mathcal{D}, \text{use\_lora}=\text{True})$\;

\Return{$\bm{\pi}_{\theta^*}$}
\end{algorithm*}

\begin{algorithm*}[t]
\caption{Sentence-Level Alignment-Enforced Translation}
\label{alg:recursive_translation}
\SetKwInput{KwInput}{Input}
\SetKwInput{KwOutput}{Output}
\SetKwFunction{FTrans}{RecursiveTranslate}

\KwInput{Source document sentences $S_\text{source} = [s_1, s_2, \dots, s_n]$}
\KwOutput{Translated target sentences $T_\text{final}$}

\SetKwProg{Fn}{Function}{:}{}
\Fn{\FTrans{$S$}}{
    \tcp{1. Construct Prompt with Indices and Separators}
    $P \leftarrow \text{EmptyString}$\;
    \ForEach{$s_i \in S$ with index $i$}{
        $P \leftarrow P + \text{``\#} i \text{ <s>} + s_i + \text{</s>''}$\;
    }
    
    \tcp{2. Perform Translation via LLM}
    $R_\text{raw} \leftarrow \text{LLM}(P)$\;
    
    \tcp{3. Parse Output and Check Alignment}
    $T \leftarrow \text{ExtractSentences}(R_\textup{raw})$\;
    
    \If{\textup{Aligned}(T, S) \textbf{or} $|S| == 1$}{
        \Return $T$\;
    }
    \Else{
        \tcp{4. Handle Misalignment via Recursion}
        
        $m \leftarrow \lfloor |S| / 2 \rfloor$\;
        $S_\text{top} \leftarrow S[1 \dots m]$\;
        $S_\text{bottom} \leftarrow S[m+1 \dots |S|]$\;
        
        $T_\text{top} \leftarrow$ \FTrans{$S_\textup{top}$}\;
        $T_\text{bottom} \leftarrow$ \FTrans{$S_\textup{bottom}$}\;
        
        \Return $T_\text{top} + T_\text{bottom}$\;
    }
}

\textbf{Main:} $T_\text{final} \leftarrow $ \FTrans{$S_\textup{source}$}\;
\Return{$T_\textup{final}$}
\end{algorithm*}

\onecolumn
\small

\begin{longtable}{p{1.15cm}p{13.8cm}}
\toprule
\multicolumn{2}{c}{\textbf{Data \& Baseline Translation}}\\
\midrule
\multicolumn{1}{c}{\multirow{10}{*}{Source}} & \begin{CJK}{UTF8}{gbsn}``肖上尉让我们搭车？''尼米奇有些惊讶地问道。``是的。说实话，他没我想象的那么糟糕。我的意思是，他虽然不怎么样，但也不至于让人讨厌。''科伦抬起拇指和食指，按着喉部的麦克风，同时将背包移到背上。``Exodus连所有单位，完成任务后到西区车辆段集合。我们将在05:00与第一装甲排一起出发。''每隔几秒钟，他就会收到班长们的回复。他们刚来的时候，哈里斯、克雷斯和拉维克都是级别较低的军士。自从在地球上进行基础训练以来，他们四人就一直是好朋友。科伦觉得认识他们仿佛是很久以前的事了，尽管这才过去不到两年。他们都是班长，但自从一周前，前排长被Daybreak狙击手击中，因伤退役后，克雷斯就开始领导第二排了。走到车辆段时，科伦和尼米奇的目光落在了大约八辆克罗诺斯主战坦克和四辆史赛克装甲车上。机组人员四处奔走，确保他们的战争机器一切正常。科伦很快就在人群中找到了博拉丁·肖上尉，他正在和装甲排的指挥官谈话。他们走上前去时，肖上尉看了他们一眼，也迎了上来，他那张惯常冷漠的脸上看不出任何情绪。\end{CJK} \\
\hdashline
\multicolumn{1}{c}{\multirow{12}{*}{Reference}} & ``Shaw got us a ride?'' Nemic asked, a little surprised. ``Yup. Honestly, he isn't as bad as I thought he was. I mean, he's not great, but he isn't unbearable'' Cohren lifted his thumb and index finger to his throat mic, shifting his pack on his back while he did. ``All units in Exodus Company, meet at Motor Pool West when you're finished. We are heading out with the 1st Armored Platoon at 0500'' Every few seconds he got a response from the squad leaders. Haris, Craith, and Ravik had all been low ranking NCO when they all got there. All four of them had been good friends since basic back on Earth. It felt so long ago now since Cohren had met them, even though it couldn't have been two years ago. They were squad leaders, But Craith was now leading 2nd Platoon as of a week ago, after the previous leader had gotten medically discharged after catching a bullet from a Daybreak sniper. Walking up to the motor pool, Cohren and Nemic's eye fell on a group of around eight Kronos Main Battle Tanks and four Strykers. Their crew milled about, making sure their war machines were ready to go. Cohren's gaze soon found Captain Boradin Shaw, who was talking with the armored platoon's commander. As they walked up, Shaw looked over and walked up, his usually cold face masking all emotions. \\
\hdashline
\multirow{14}{*}{\makecell{Baseline\\Translation\\sCMT: 81.1}} & ``Did \textbf{Lieutenant} Xiao let us ride with him?'' \textbf{Nimitz} asked in surprise. ``Yes. To be honest, he's not as bad as I imagined. I mean, he's not great, but he's not\begin{CJK}{UTF8}{gbsn}讨厌\end{CJK} either.'' \textbf{Cole} raised his thumb and index finger, pressing against the microphone on his throat, while shifting his backpack onto his back. ``All units of Exodus, assemble at the west district vehicle depot after completing the mission. We will depart with the first armor platoon at 05:00.'' Every few seconds, he would receive replies from the squad leaders. When they first arrived, Harris, \textbf{Cress}, and Ravek were all junior non-commissioned officers. Since their basic training on Earth, the four of them have been close friends. \textbf{Cole} felt like he had known them for a long time, even though it had been less than two years. They were all squad leaders, but since a week ago, when \textbf{the front sergeant} was shot by Daybreak snipers and had to retire due to injuries, \textbf{Cress} began leading the second squad. As they approached the vehicle depot, \textbf{Colonel Kolen} and \textbf{Admiral Nimitz}'s eyes fell upon roughly eight Cronos main battle tanks and four Stryker armored vehicles. Crew members were bustling around, ensuring their war machines were all functioning properly. \textbf{Kolen} quickly spotted \textbf{Captain Borlatan Xiao} in the crowd, who was engaged in conversation with the commander of the armored platoon. As they approached, Captain Xiao glanced at them and came forward, his usual indifferent expression revealing no emotion. \\
\bottomrule
\caption{A case of source and reference segment with the translation results of the baseline method. ``sCMT'' denotes the sCOMET score.}\label{tab:case_baseline}
\end{longtable}

\begin{longtable}{p{1.15cm}p{13.8cm}}
\toprule
\multicolumn{2}{c}{\textbf{\dragon Translation Process}}\\
\midrule
\multicolumn{1}{c}{\multirow{6}{*}{Essence}}
& [1] \begin{CJK}{UTF8}{gbsn}科伦和尼米奇被派往伊安洛斯城外支援地面进攻，采用``野鼬鼠''战术对抗敌方轰炸机，前线战事紧张，他们将与第一装甲排一同行动。\end{CJK}\\
& [2] \begin{CJK}{UTF8}{gbsn}科伦作为中尉，带领Exodus连准备行动，成员们迅速整理装备，15分钟内完成检查，准备出发。\end{CJK}\\
& [3] \begin{CJK}{UTF8}{gbsn}科伦回应问题，门突然打开，哈里斯、克雷斯等人已穿戴整齐，准备行动，克雷斯询问是否被什么吵醒。\end{CJK}\\
& [4] \begin{CJK}{UTF8}{gbsn}科伦按响警报叫醒士兵，拉维克抱怨早餐糟糕，新兵们虽困倦但已穿戴整齐准备就绪。\end{CJK}\\
\hdashline
\multicolumn{1}{c}{\multirow{12}{*}{Analysis}}
& [1] This summary provides context about the mission and the tactical approach (``wild goose'' tactic), but it is not directly relevant to the current passage, which focuses on the coordination and preparation for the mission rather than the overall strategy. Therefore, it should not be selected.\\
& [2] This summary mentions the preparation of the Exodus Company and the time frame for gear checks, but the current passage is more about the communication and the relationship dynamics among the soldiers rather than the timeline of preparation. Therefore, it should not be selected.\\
& [3] This summary refers to the soldiers being awakened and the interaction with Krass, but the current passage does not involve waking up or direct dialogue with Krass in the same context. Therefore, it should not be selected.\\
& [4] This summary includes details about the soldiers being awakened, the mention of Lavik complaining about breakfast, and the overall readiness of the troops. These details are relevant to the current passage, which includes the soldiers being called to action and their state of readiness. Therefore, it should be selected.\\
\hdashline
\multicolumn{1}{c}{Selection} & 4 \\
\midrule
\multicolumn{1}{c}{\multirow{19}{*}{Exemplar}}
& [1] [Source] \begin{CJK}{UTF8}{gbsn}伊塔尼亚时间，凌晨 4:30。科伦讨厌这么早起床。主要是因为他要负责叫醒连队的其他人。他默默地走下床，穿上作战服，套上突击背心，从储物柜里收拾好东西，小心翼翼地拿起一个特别的小盒子。\end{CJK} // [Target] Itania time, 4:30 a.m. Korren hated waking up so early. Mainly because he was responsible for waking up the rest of the squad. He quietly got out of bed, put on his combat gear, strapped on his assault vest, packed his belongings from the storage cabinet, and carefully picked up a small special box.\\
& [2] [Source] \begin{CJK}{UTF8}{gbsn}他从夹克口袋里掏出打火机，又从另一个口袋里掏出一支烟。点燃香烟后，他轻轻地吸了一口，然后打开了门。\end{CJK} // [Target] He took out a lighter from his jacket pocket and a cigarette from another pocket. After lighting the cigarette, he took a gentle puff and then opened the door.\\
& [3] [Source] \begin{CJK}{UTF8}{gbsn}``伊安洛斯城外的某个前线基地。我们被派去支援地面进攻。''科伦说着，吸了一口烟。他有点儿责怪克雷斯让他染上了烟瘾。\end{CJK} // [Target] ``Some frontline base outside of Ianlos. We've been sent to support the ground assault.'' Korren said, taking a puff of his cigarette. He somewhat blamed Krass for getting him addicted to smoking.\\
& [4] [Source] \begin{CJK}{UTF8}{gbsn}``带上你们的私人物品，''科伦继续说道。作为一名中尉和连队的副指挥官，他有责任协助连长博拉丁·肖上尉监督和领导连队。``我们不会回来了。各位军士和少尉，给你们和你们的队员15分钟时间做最后的检查，并带上今天所需的口粮。''\end{CJK} // [Target] ``Pack your personal belongings,'' Korren continued. As a \textbf{lieutenant} and deputy commander of the squad, he was responsible for assisting \textbf{Captain Borlatin Xiao} in supervising and leading the squad. ``We won't be coming back. Sergeants and lieutenants, you and your squads have 15 minutes to do your final checks and bring along the rations needed for today.''\\
\hdashline
\multicolumn{1}{c}{\multirow{14}{*}{Analysis}}
& [1] This sentence pair provides detailed description of Korren's morning routine and preparation for the mission. However, the current passage is focused on the coordination and communication during the pre-mission setup, not on the individual preparation process. Therefore, it should not be selected.\\
& [2] This sentence pair includes details about Korren's method of waking up his subordinates and his frustration with the alarm system. While it offers insight into his leadership style, the current passage is about the coordination and readiness of the squad rather than the process of waking up. Therefore, it should not be selected.\\
& [3] This sentence pair provides context about the mission objective (``support the ground assault'') and includes a dialogue about smoking, which is relevant to understanding the character dynamics and the mission context. The mention of ``smoking'' also helps in understanding the tone and relationship between Korren and Krass. Therefore, it should be selected.\\
& [4] This sentence pair includes key information about the mission structure, Korren's role as a deputy commander, and the timeline for final checks. It directly relates to the current passage, which discusses the coordination and instructions given to the squad before the mission. Therefore, it should be selected.\\
\hdashline
\multicolumn{1}{c}{Selection} & 3, 4 \\
\midrule
\multicolumn{1}{c}{\multirow{23}{*}{Entity}}
& [1] \begin{CJK}{UTF8}{gbsn}科伦\end{CJK} / Korren: A lieutenant who is responsible for assisting the company commander Captain Borlatin Xiao in leading the squad, wearing combat gear and carrying weapons, demonstrating a sense of responsibility for the mission.\\
& [2] \begin{CJK}{UTF8}{gbsn}哈里斯\end{CJK} / Harris: Harris is a soldier in Korren's unit and is one of the troops Korren is responsible for waking up, showing preparedness and readiness during the early morning routine.\\
& [3] \begin{CJK}{UTF8}{gbsn}克雷斯\end{CJK} / Krass: Krass is Korren's comrade, a soldier already dressed and ready, displaying strong self-discipline and a sense of responsibility, capable of quickly preparing equipment.\\
& [4] \begin{CJK}{UTF8}{gbsn}拉维克\end{CJK} / Lavik: Lavik is a soldier who dislikes early mornings and strict training but still follows orders, interacting with other soldiers like Korren, Harris, and Krass.\\
& [5] \begin{CJK}{UTF8}{gbsn}博拉丁·肖上尉\end{CJK} / Captain Borlatin Xiao: Captain Borlatin Xiao is the company commander of Korren's unit, responsible for leading the Exodus Squad to complete missions, maintaining a direct superior-subordinate relationship with Korren.\\
& [6] Exodus\begin{CJK}{UTF8}{gbsn}连\end{CJK} / Exodus Squad: Exodus Squad is an elite unit affiliated with the Second Armored Division, responsible for combat operations and ground assault support, using Daybreak tactics.\\
& [7] \begin{CJK}{UTF8}{gbsn}尼米奇\end{CJK} / Nemychev: Nemychev is Korren's superior lieutenant, with a pessimistic personality and dissatisfaction with war, but still performs his duties and acts alongside Korren cooperatively.\\
& [8] Daybreak / Daybreak: Daybreak is a specialized and elite military unit within the Second Army's Second Assault Division, known for its advanced ``wild boar'' tactics and combat readiness.\\
& [9] \begin{CJK}{UTF8}{gbsn}第一装甲排\end{CJK} / First armored squad: First armored squad is the supporting unit of Korren's company, which will act together with them and provide mobile support, with Captain Xiao appearing at the assembly point.\\
& [10] \begin{CJK}{UTF8}{gbsn}肖上尉\end{CJK} / Captain Xiao: Captain Xiao is Korren's superior, the company commander of the Exodus Squad, responsible for leading the entire company and ensuring the smooth completion of missions.\\
\hdashline
\multicolumn{1}{c}{\multirow{18}{*}{Analysis}}
& [1] This record provides a general description of Korren's role and habits, but it is too broad and does not offer specific contextual details relevant to the current passage. Therefore, it should not be selected.\\
& [2] This record gives information about Harris's role in the unit and his preparedness, but the current passage does not directly mention Harris or focus on his actions. Therefore, it should not be selected.\\
& [3] This record describes Krass's preparedness and discipline, which is relevant to the context of the squad being ready, but the current passage does not focus on Krass or his interactions. Therefore, it should not be selected.\\
& [4] This record provides information about Lavik's personality and role, but he is a background character and not directly involved in the current scene. Therefore, it should not be selected.\\
& [5] This record explains Captain Borlatin Xiao's role as the leader of the Exodus Squad and his relationship with Korren. This is directly relevant to the current passage, where they are coordinating and preparing for the mission. Therefore, it should be selected.\\
& [6] This record provides general information about the Exodus Squad, but the current passage focuses on specific interactions and instructions rather than the broader unit structure. Therefore, it should not be selected.\\
& [7] This record describes Nemychev's role and personality, but he is not a central character in the current scene. Therefore, it should not be selected.\\
& [8] This record gives background on the Daybreak unit and its tactics, but the current passage does not mention Daybreak or its tactics. Therefore, it should not be selected.\\
& [9] This record explains the role of the First Armored Squad, which is mentioned in the current passage as being part of the mission. It helps in understanding the coordination and structure of the operation. Therefore, it should be selected.\\
& [10] This record is essentially a duplicate of Record 5, as it refers to the same person (Captain Xiao) and his role. Since Record 5 is already selected, this record is redundant. Therefore, it should not be selected.\\
\hdashline
\multicolumn{1}{c}{Selection} & 5, 9\\
\midrule
\multirow{13}{*}{\makecell{\dragon \\ Translation\\sCMT: 82.4}}
& ``\textbf{Captain} Xiao let us ride?'' \textbf{Nemic} asked, surprised. ``Yes. To be honest, he's not as bad as I imagined. I mean, he's not great, but he's not\begin{CJK}{UTF8}{gbsn}讨厌\end{CJK} either.'' \textbf{Korren} raised his thumb and index finger, pressing against his throat's microphone, while he moved his backpack onto his back. ``All units of Exodus Squad, assemble at the West Zone Vehicle Yard after mission completion. We will depart with the First Armored Squad at 05:00.'' Every few seconds, he received replies from the squad leaders. When they first came, Harris, \textbf{Krass}, and Ravek were all lower-ranking sergeants. Since basic training on Earth, the four of them have been close friends. \textbf{Korren} felt like he had known them for a long time, even though it had been less than two years. They were all squad leaders, but since a week ago, when \textbf{the previous squad leader} was shot by Daybreak snipers and retired due to injury, \textbf{Krass} started leading the Second Squad. When they arrived at the vehicle yard, \textbf{Korren and Nemic}'s eyes fell on about eight Cronos main battle tanks and four Stryker armored vehicles. Crew members were running around, ensuring their war machines were in working order. \textbf{Korren} quickly found \textbf{Captain Borlatin Xiao} in the crowd, who was talking to the commander of the armored squad. As they approached, Xiao looked at them and came forward, his usual indifferent face showing no emotion. \\
\bottomrule
\caption{A case of the translation process of \dragon. ``sCMT'' denotes the sCOMET score.}\label{tab:case_ours}
\end{longtable}

\end{document}